\documentclass[12pt]{book}
\pdfoutput=1
\usepackage[english]{babel}
\usepackage[utf8x]{inputenc}
\usepackage{amsmath}
\usepackage{graphicx}
\usepackage{authblk}
\usepackage[numbers]{natbib}
\graphicspath{{Images/}}
\usepackage[colorinlistoftodos]{todonotes}
\usepackage{hyperref}
\usepackage{multirow}
\usepackage{pdfpages}
\usepackage{textcomp}
\usepackage{graphicx}
\usepackage{subfig}

\usepackage{tikz}
\usetikzlibrary{shapes,arrows}

\usepackage{longtable}

\tikzstyle{decision} = [diamond, draw, fill=blue!20,
    text width=7.5em, text badly centered, node distance=3cm, inner sep=0pt]
\tikzstyle{block} = [rectangle, draw, fill=blue!20,
    text width=10em, text centered, rounded corners, minimum height=4em]
\tikzstyle{block-horizontal} = [rectangle, draw, fill=blue!20,
    text width=4em, text centered, rounded corners, minimum height=4em]
\tikzstyle{block-dashed} = [rectangle, draw, fill=blue!20, dashed,
    text width=15em, text centered, rounded corners, minimum height=4em]
\tikzstyle{line} = [draw, -latex']
\tikzstyle{cloud} = [draw, ellipse,fill=red!20, node distance=1cm,
    minimum height=2em]

\begin{document}
\begin{titlepage}
\newcommand{\HRule}{\rule{\linewidth}{0.1mm}} 
\center 
 
\textsc{\Large Monitoring spatial sustainable development:}\\[0.5cm] 
\textsc{\Large Semi-automated analysis of satellite and aerial images for energy transition and sustainability indicators}\\[0.5cm] %
\HRule \\[0.4cm]
{ \bfseries Eurostat Grant Agreement: 08143.2017.001-2017-408}\\[0.1cm] 

---------------------------------------------------------------------------------

\HRule \\[0.4cm]
{ \huge \bfseries Final Report }\\[0.1cm] 
\HRule \\[1.5cm]

{\large Date of issue : \today}\\[1cm] 
\vfill 
 
\newpage
\begin{table}
\centering
\begin{tabular}{|l|l|}
\hline
& Name\\
\hline
\multirow{ 4}{*}{Author}& De Jong, T.J.A. (CBS) \\
	        & Bromuri, S (OU)\\
	        & Chang, Xi (OU) \\
	        & Debusschere, M. (Statbel) \\
	        & Rosenski, N. (Destatis) \\
	        & Schartner, C (Destatis) \\
	        & Strauch, K (IT.NRW) \\
	        & Boehmer, M. (IT.NRW) \\
	        & Curier, R.L. (CBS) \\

\hline
Reviewed by & 	 \\
\hline
Distribution List & 	 \\
\hline
\end{tabular}
\end{table}
\HRule \\[1.5cm]
\begin{table}
\caption*{Document change records}
\begin{tabular}{|l|l|l|l|}
\hline
Date & Issue & Affected & Description\\
\hline
14-01-2019 & Draft&All& First draft\\
\hline
21-08-2019 & Draft&All& Second draft\\
\hline
22-10-2019 & Draft&All& Final version ready for submission\\
\hline
\end{tabular}
\end{table}

\end{titlepage}

\tableofcontents        
\newpage

\chapter*{Executive Summary}
This report presents the results of the DeepSolaris project that was carried out under the ESS action 'Merging Geostatistics and Geospatial Information in Member States'. The project was carried out by a consortium of four NSOs and one university, spread across three countries: the Netherlands, Germany, and Belgium. For the Netherlands, Statistics Netherlands contributed by delivering geospatial information about the Netherlands as well as developing the deep learning algorithms together with the Open Universiteit Nederland. For Germany, IT.NRW and Destatis contributed geospatial information and knowledge about North Rhine Westfalia in Germany. Likewise, for Belgium, Statbel contributed the geospatial information and knowledge. 

During the project several deep learning algorithms were evaluated to detect solar panels in remote sensing data. The aim of the project was to evaluate whether deep learning models could be developed, that worked across different member states in the European Union. Two remote sensing data sources were considered: aerial images on the one hand, and satellite images on the other. It was soon found that the resolution of the satellite images available (from the Pleiades and Superview satellites) was insufficient for the task at hand. After that, the project focuses on aerial images, in specific on high resolution images with a resolution of 10cm/pixel. These aerial images were only available for the province of Limburg in the Netherlands and the state of North Rhine Westfalia in Germany. 

Two flavours of deep learning models were evaluated: classification models and object detection models. Classification models give a coarse overview whether a certain area contains solar panels or not. Object detection models can distinguish the location of individual solar panels and can help with detailed statistics about the number of solar panels, the area, and the expected yield. Both models were trained by a process called supervised learning in which a dataset contains the images to train the algorithm and labels for each image containing the so-called ground truth. For the classification task this entails labeling each image with a binary label: image contains solar panels versus does not contain solar panels. For an object-detection algorithm, this entails drawing the boundaries of each individual solar panel. As such, the labeling process for the object detection task is a lot more labour intensive than for the classification task where a simple yes/no for the entire image suffices. For this reason, the object detection model was only evaluated for the North Rhine Westfalia region on a much smaller dataset that those used for the classification task. 

For the evaluation of the deep learning models we used a cross-site evaluation approach: the deep learning models where trained in one geographical area and then evaluated on a different geographical area, previously unseen by the algorithm. The cross-site evaluation was furthermore carried out twice: once on the same aerial image in a different geographical area in the same country, and once in a cross-border fashion on a different aerial image in a different country; deep learning models trained on the province of Limburg in the Netherlands were evaluated on the state of North Rhine Westfalia, Germany and vice versa. While the deep learning models were able to detect solar panels successfully, false detection remained a problem. In this case, for example cars, greenhouses, and rooftop windows were detected as if they were solar panels. It was also found that model performance differed between urban and more rural areas in the same geographical region. Moreover, model performance decreased dramatically when evaluated in a cross-border fashion. Hence, training a model that performs reliably across different countries in the European Union is a challenging task. That being said, the models detected quite a share of solar panels not present in current solar panel registers and therefore can already be used as-is to help reduced manual labor in checking these registers. Several directions for future work have been identified, that could aim to improve model performance by trying to understand the causes of deteriorating model performance and trying to deal with them.

\chapter*{Abstract}
%

\textbf{This report describes the results of applying Deep Learning algorithms to aerial images in the regions of Limburg (Netherlands) and to aerial and satellite images in North Rhine Westphalia (Germany). The report highlights that the selected methods manage to create algorithms that are predictive with respect to the task, but it also highlights how the false positive rate is an issue difficult to overcome. The main result of the project is that it is possible to train deep learning algorithms to detect solar panels in aerial images, but the false positive/negative rates do not allow to make it a fully automated service. The algorithms developed could though be used to improve the solar panels registry as they manage to identify solar panels that were not previously known by the Dutch and German partners.}

\chapter{Introduction}

In recent years, National Statistic Offices (NSOs) have increasingly searched for ways to use new data sources. Traditionally, most official statistics have been based on information retrieved via surveys. While still relevant, surveys also have their range of problems. More and more non-response to surveys is becoming a problem, especially in certain groups of society. Surveys also take a lot of time to create, carry out, and to process, making it difficult, if not impossible, to create (near) real-time statistics. In addition, they only reflect a sample of society, which may be subject to certain biases, especially when certain sub populations are under- or over-represented. It is therefore that NSOs have looked into alternative data sources that replace survey (questions) altogether or paint a different part of the picture and serve as an additional form of information. Several of such data sources can be identified. Some of these data sources, the administrative data sources, have been collected by governmental institutions and give a unique and often integral view of society. While often not directly collected for the purpose of official statistics, administrative data can often be used and processed in similar ways to survey data and without using special techniques. Other data sources, however, like for example big data and sensor data, are not collected with official statistics in mind and need special techniques to be processed and refined into official statistics. Although often much more difficult to process, these data sources can often give us an extra dimension to existing data sources, can give a real-time or integral view on society, or can provide completely new information absent from any survey or administrative data source.

In this report, we will highlight one of these new data sources: images, and in specific aerial and satellite images. The report will focus on using deep learning algorithms to retrieve information from images and the use of this information for the production of official statistics. In specific, we want to automate the process to extract the location of solar panels from aerial or satellite images and produce a map of solar panels along with regional statistics on the number of solar panels. The detection of solar panels is only one possible application which can show the potential and benefits of deep-learning algorithms in official statistics. At the outset of the project we had several research questions:
\begin{enumerate}
    \item What is the usability of different types of images (i.e. aerial and satellite) and minimally required image resolution for detecting solar panels?
    \item Which method is best suited for detecting solar panels?
    \item Is it possible to develop a harmonized method across EU member states?
\end{enumerate}

In the following chapters, we will address these three research questions in detail. 

The structure of this report is as follows. The methodology, datasets, and machine learning algorithms used are presented  \ref{chapter:methodology}. Chapter \ref{chapter:results} presents the results of  the different phases and  various case studies which have been carried out during the project.  Conclusions and  suggestions for future work are discussed in chapter \ref{chapter:outlook} and \ref{chapter:further_work} respectively.

\chapter{Methodology}
\label{chapter:methodology}
To automate the process of solar panel detection in aerial or satellite images several aspects need to be considered. First, possible remote sensing data sources need to be identified and assessed for their suitability for the task (see section \ref{section:datasets}). Second, relevant deep learning methods and algorithms need to be evaluated (see section \ref{section:deep_learning_algorithms}). Third, to train a deep learning algorithm, the data needs to be labeled and software to carry out this labeling needs to be developed (see \ref{section:data_annotation}). Fourth, the chosen deep learning algorithm need to be trained and tested (see section \ref{section:model_training}. We trained several models on two distinct geographical regions: the state of North Rhine Westphalia in Germany and the province of Limburg in Netherlands. Fifth, the deep learning algorithm needs to be validated (see section \ref{section:validation}). To this cause, we tested models trained on one geographical area and evaluated the trained models on the other area. Furthermore, a comparison of the algorithm results with solar panel registries can be used to validate the algorithm as well as the registers. Last, we conclude this chapter with an overview of related work in section \ref{section:recent_work}.


\section{Datasets}
\label{section:datasets}

Aerial images are images that are typically taken from a flying object, such as an airplane, helicopter, or unmanned drone\footnote{\url{https://en.wikipedia.org/wiki/Aerial_photography}}. Aerial images are taken from great height and provide a top-down view of a geographical area that can be used for various purposes, such as detailed overview of the topography of the geographical area or an overview of land use. Three datasets were available but only two datasets were used:
\begin{enumerate}
    \item  Aerial images for \emph{North Rhine Westphalia Data (NRW)} are available from OpenGeodata. The NRW data are available at 10x10 cm. We collected 50000 images from NRW containing solar panels, using the LANUV registry. 5000 of these images were further annotated with the polygon containing the solar panel. Each of the images present a dimension of 330x330 pixels, to be compatible with most deep learning frameworks. Such images were used to train two algorithms: one classification algorithm (using the 50000 images with solar panels and 50000 images without) and an object detection algorithm (using only the polygon annotations).
    
    \item For the Netherlands, each year two aerial images are taken in commission of the Dutch Government: a low resolution image (25cm per pixel) in the beginning of spring and the summer, and a high resolution image (10cm per pixel) in the leafless season. Images are available in both 24-bit RGB as well as CIR (infrared). Typically a campaign to create an aerial image of the Netherlands takes several months. As an example, for 2019, the low resolution image was taken in the period from February until June\footnote{\url{http://www.beeldmateriaal.nl/index.html}}. Statistic Netherlands receives the low resolution images (25cm per pixel) on a yearly basis and therefore several years of aerial images since 1995 are present in its archive. In addition, since 2016 the low resolution images are also made available on a yearly basis on the PDOK website\footnote{\url{https://www.pdok.nl/}}. Images from the region of Limburg are also available in a higher resolution. Specifically, the images from Limburg were collected in 2018 and present a 10x10 cm level of detail, like the NRW images. In addition to the aerial image dataset, we used the Dutch building register (BAG) to filter out those images that contained buildings, largely reducing our final dataset. In addition, we used the Statistic Netherlands' register on solar panel locations to make a preliminary selection of the buildings containing solar panels.
    
    \item For the region of Flanders in Belgium, images are available in DOP25 format. During the evaluation of our models (see Section \ref{sec:results}), we noticed that the transfer of a model from a region to another already results in a significant performance drop. Furthermore, the two regions evaluated (NRW and Limburg), were both available in the higher-quality DOP10 format. Because we expected an even further performance drop when using the lower resolution DOP25 image for Flanders, we decided to focus on NRW and Limburg.
    
\end{enumerate}

For each of the datasets, two separate sets of images were created. On on the one hand, a $200\times200$ pixel or $20\times20m$ dataset was created and annotated to train a neural network for the province of Limburg, the Netherlands. This algorithm was evaluated on a dataset of the same resolution taken from NRW. On the other hand, a $330\times330$ or $33\times33m$ dataset was created to train algorithms above NRW. Similarly, a dataset of an equivalent resolution was created to evaluate NRW network performance above Limburg.

\subsection{Satellite Images}
\label{section:phase5_satellite_images}

The case making use of the satellite had a few hurdles in terms of data availability and access to a perennial source of data. Currently the satellite observation which are available free of charge do not allow for the detection of solar panels on single dwellings as the spatial resolution is too low.

A small subset of  Pleiades images was acquired by Destatis for a study area of 450 km²  which covers a very heterogeneous area regarding rural and urban areas and with it various housing structures at 50x50cm$^2$ resolution. Similarly, Superview images for Heerlen with a resolution of 50x50cm$^2$ resolution were acquired via the The Netherlands Space Office (NSO) satellite portal. The access to some commercial satellite data is facilitated as NSO provides access to satellite data from the Netherlands to Dutch users.

In practice, the remote sensing image was chosen based on the criteria:

\begin{itemize}
	\item The cloud coverage of the satellite images should be minimal. A meaningful analysis is only possible when no clouds or cloud shadows obscure the image.
\item The recording date of the image should be no later than December 2017 to match the currency of the administrative data of LANUV. Ideally, the images should be from the years 2016 to 2017 to match the aerial images.
\item Furthermore, the satellite images should at least roughly match the test area used in the aerial images. 
\end{itemize}

\begin{figure}[h]
  \includegraphics{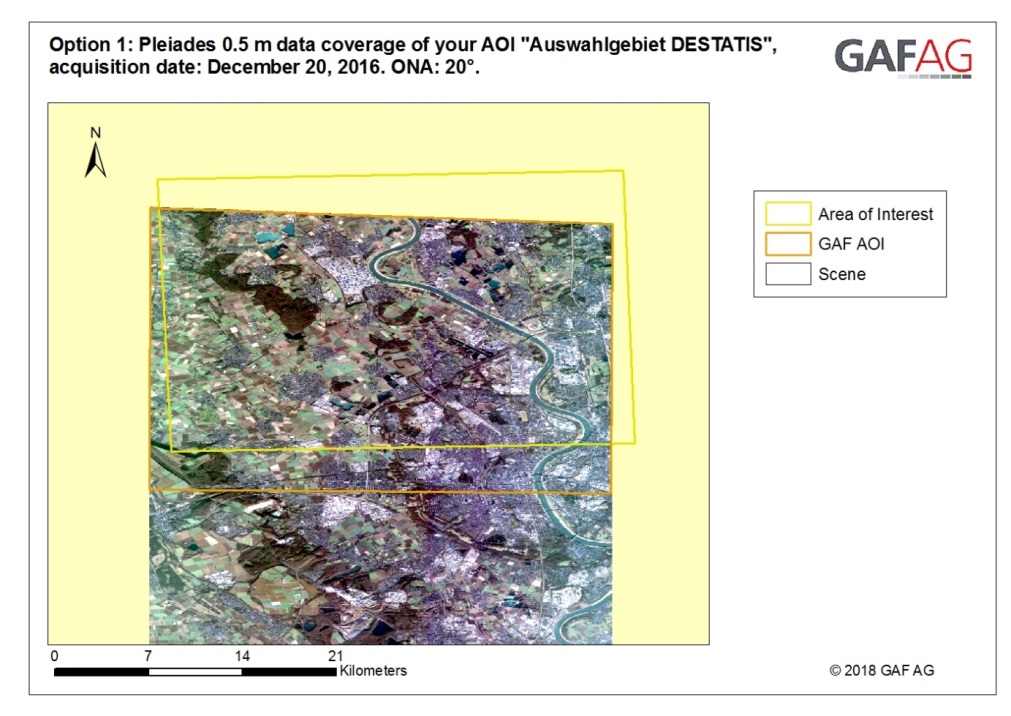}
  \caption{Pleiades image}
  \label{fig:plei}
\end{figure}

\begin{figure}[h!]
  \includegraphics[scale=0.4]{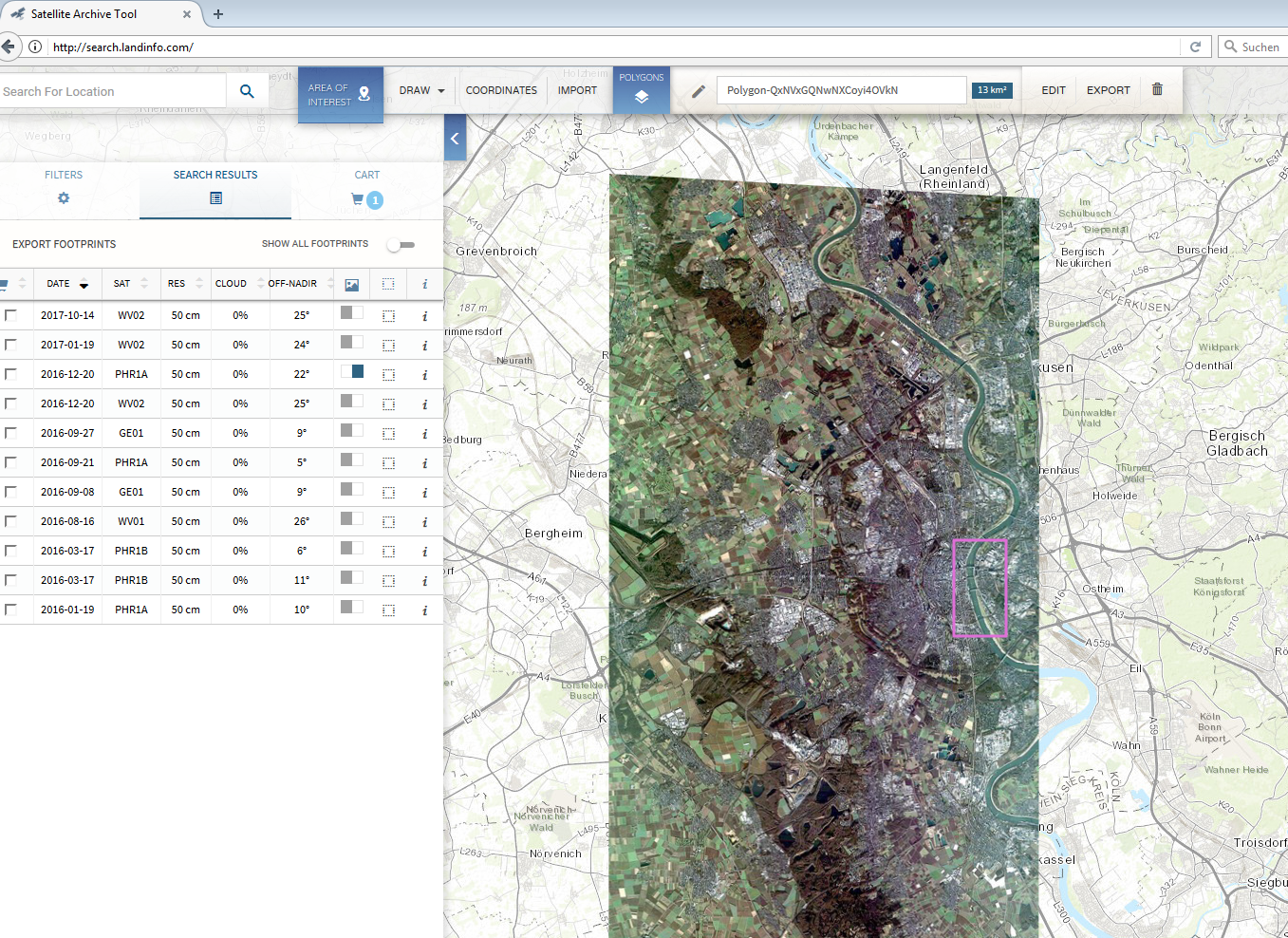}
  \caption{Availability of cloud free very high resolution images}
  \label{fig:avail}
\end{figure}

The remote sensing image chosen based on the aforementioned requirements was taken on December 20th, 2016 by a Pleiades Satellite as shown in figure \ref{fig:plei}. The Pleiades sensor has a resolution of 50 cm panchromatic and 2 meter multi-spectral. The image was however taken at an angle of 20 degrees, which means the resolution is somewhat lower. A higher resolution than 50cm was not available for the area of interest. Figure \ref{fig:avail} shows the search results for the restrictions of area of interest, chosen time period and limited cloud coverage and the chosen image. The images which are available for this scene have a resolution of 50cm. 

\begin{figure}[h!]
  \includegraphics[scale=0.7]{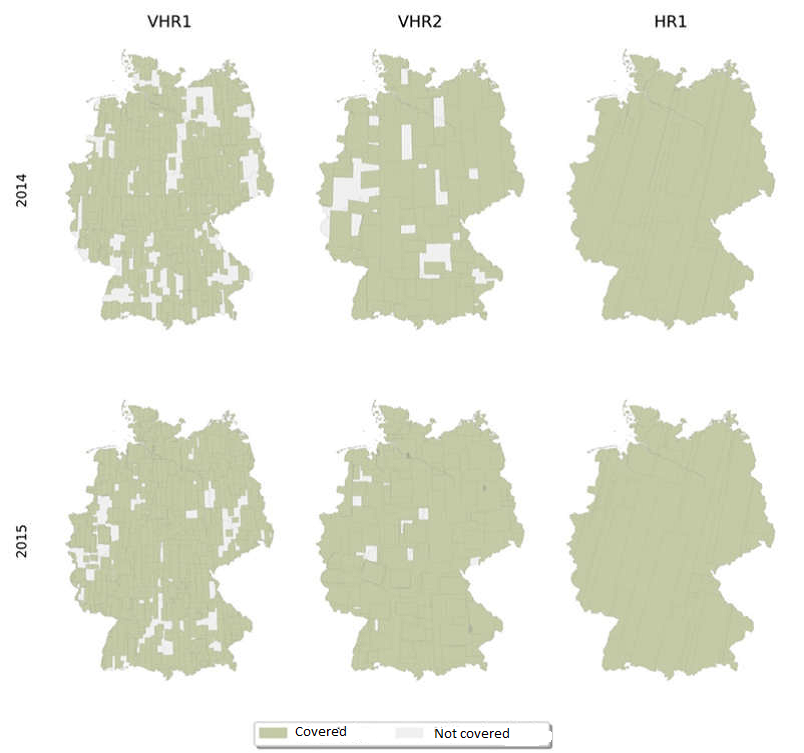}
  \caption{Coverage of high resolution satellite images: VHR1 (resolution$<$1m), VHR2 (resolution 1-4m) and HR1 (resolution 4-10m) \citep{ZKI} }
  \label{fig:cov}
\end{figure}

Overall the availability of very high resolution images is low or infrequent as figure \ref{fig:cov} shows. The swath width of very high resolution images is very small, which means an image does not cover a large area. The very high resolution images are only taken if they are ordered by a client, which means that the coverage between areas varies greatly.

\paragraph{The case study on satellite images can not be investigated} In view of the current quality and resolution of satellite data available to us it was evaluated that satellite data do not provide a viable and perennial source of data which could be used to automate the detection of solar panel  by means of machine learning. In practice, data from Pleaides (pixel size 50cm), Superview (pixel size 50cm) and Sentinel-2 (pixel size 10m) was evaluated. For all instruments the detection of solar panel could not go further as the solar panel could not be seen by the annotator, indeed, even when knowing the location of the solar panel they could not be identified on the images and therefore a training set could not be created. On average a PV solar panel is 99x165cm$^2$ and 99x198cm$^2$ for residential and commercial use respectively which means that, in case of Pleiades and Superview a solar panel spans over an area of 3x2 pixels in average while it is lost somewhere inside a single Sentinel-2 pixel.

Hence, in the remainder of the document the methodology and results presented in this paper solely focus on the suitability of aerial images to detect solar panels.

\section{Deep Learning Algorithms}
\label{section:deep_learning_algorithms}

The recent popularity of deep learning started in 2012 when the seminal AlexNet \citep{Krizhevsky:2012:ICD:2999134.2999257} outperformed all other classifiers on the ImageNet Large Scale Visual Recognition Challenge (ILSVRC)\citep{ILSVRC15} by a large margin (11\% to the second-best entry). In the ILSVRC, computer vision algorithms are presented with a classification challenge in which they need to distinguish a 1000 different categories. In subsequent years, convolutional neural networks, like VGG16 \citep{Simonyan2014}, Inception \citep{Szegedy2015}, ResNet \citep{DBLP:journals/corr/HeZRS15}, and Xception \citep{DBLP:journals/corr/Chollet16a}, were used to outperform the benchmark set by AlexNet and achieve superhuman performance \citep{DBLP:journals/corr/GeirhosJSRBW17}. As was shown by AlexNet\citep{Krizhevsky:2012:ICD:2999134.2999257} and its successors, deep convolutional networks trained with stochastic gradient descent (or variations of it like RMSprop and Adam) \citep{DBLP:conf/nips/LeCunBOM96, Tieleman2012, DBLP:journals/corr/KingmaB14}, make excellent end-to-end classifiers. Moreover, convolutional neural networks are also getting more and more adept at object detection \citep{DBLP:journals/corr/GirshickDDM13, DBLP:journals/corr/Girshick15, DBLP:journals/corr/RenHG015, DBLP:journals/corr/RedmonDGF15, DBLP:journals/corr/LiuAESR15}. Within this project we studied two types of approaches with respect to the detection of solar panels:

\begin{itemize}
    \item Image Classification
    \item Object Detection 
\end{itemize}
Furthermore, we investigated pre-existing deep-learning architectures. These architectures are available in the Keras \citep{chollet2015keras} library with pre-trained weights. Below we explain the theory behind image classification, object detection, and transfer learning.

\subsection{Image Classification}

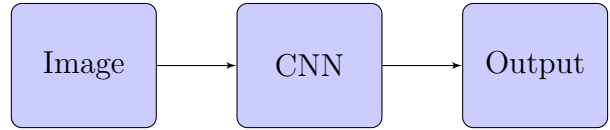
\begin{figure}[!h]
\begin{center}   
\begin{tikzpicture}[node distance = 3cm, auto, transform shape]
    \node [block-horizontal] (image) {Image};
    \node [block-horizontal, right of=image] (classifier) {CNN};
    \node [block-horizontal, right of=classifier]  (output) {Output};
    
    \path [line] (image) -- (classifier);
    \path [line] (classifier) -- (output);
\end{tikzpicture}

\caption{Deep Learning where a convolutional neural network is trained using an optimizer to learn the features it needs to perform the classification. A convolutional neural network is therefore an example of an end-to-end classifier taking raw RGB data as input and giving the class predictions as output.}
\label{fig:deep_learning_classifier}
\end{center}
\end{figure}

In image classification, the algorithm receives an image and has to tell whether the image contains a solar panel or not. As can be seen in figure \ref{fig:deep_learning_classifier},  these end-to-end classifiers can be trained with only the raw image data as input and the corresponding classification label as output. The features needed to perform the classification task are learned from the data, bypassing the need for specifying hand-crafted features. The trade-off of using convolutional neural networks is that they need a much larger amount of data as compared to the amount of data when using traditional computer vision with hand-crafted features. However, the performance of convolutional neural networks often drastically outperforms the traditional computer vision approaches.

\subsection{Object Detection}

\begin{figure}
    \centering
    \includegraphics[width=0.9\textwidth]{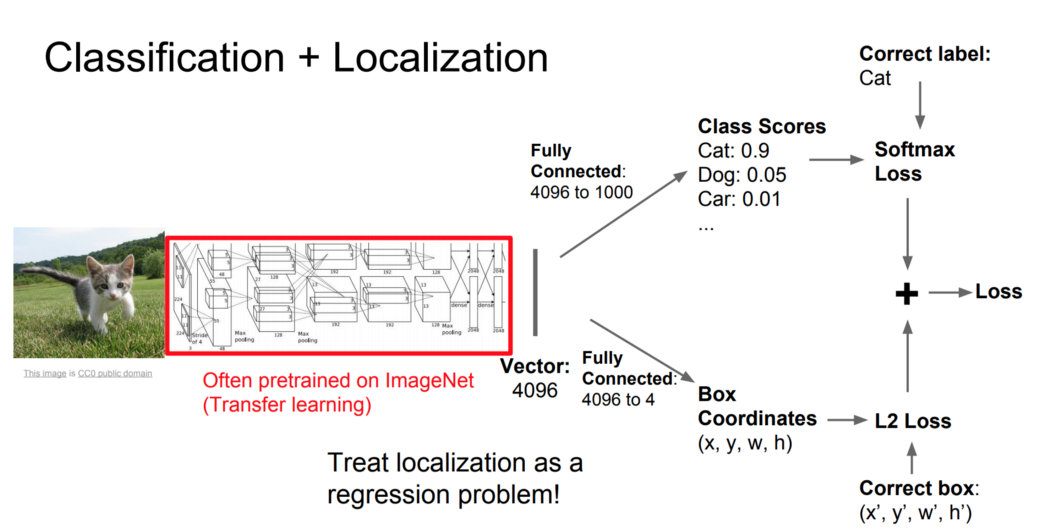}
    \caption{Object Localization}
    \label{fig:objconcept}
\end{figure}

Different from basic classification, object detection is concerned with detecting the presence of objects in an image. Various approaches exist towards object detection, like for example, using sliding windows on an image, anchors, bounding boxes or intersection over union (see \cite{cheng2016survey} for a survey on the subject). The basic concept of object detection can be understood as an extension of image classification in which, in addition to detecting that an object is present in the image, we also detect the bounding boxes of the object. The simplest approach is therefore to regress the location of a single bounding box in the image,  as shown in figure \ref{fig:objconcept}.

In the simplest case, this effectively means regressing $x,y,w,h$ parameters of the bounding box, with $x$ and $y$ being the location of the top left corner and $w$ and $h$ being the width and height of the bounding box respectively.

\subsection{Transfer Learning}
\label{sec:TL}
The training of current state-of-the-art convolutional neural networks for large datasets is both computationally expensive and time consuming. Therefore, in our case, training a CNN from scratch is not an ideal option and transfer learning is preferred. The principal advantage of a transfer learning strategy is to quickly create powerful models. Transfer learning refers to the use of existing pre-trained models to classify images on which they have not been trained and which may even be labelled with different categories. Transfer Learning can be done via various approaches:
\begin{itemize}
    \item \emph{The new data set is rather small and very different from the original data set}.  All but the last, fully-connected layers of the pre-trained model are considered and used as feature extractors as it was done in traditional Computer Vision approaches. A machine-learning classifier (e.g. a SVM) is trained on top of the layers with the labels of the new data set.
    \item \emph{The new data set is very different from the original data set but large}. This approach fine-tunes the weights of the pre-trained model, either for all layers or just for some of the later ones, while keeping the remaining layers frozen. 
    \item Finally, \emph{some of the layers of a pre-trained network are used as baseline for a new model}, followed by the typical convolutional building blocks which would be trained from scratch.
\end{itemize}

In general, earlier layers of a CNN pick up more general features, similar to Gabor filters or colour blobs, while the higher-layer neurons are specialized to specific tasks and learn increasingly more complex/composite features. 

\section{Data Annotation}

\label{section:data_annotation}

An important part in training deep learning algorithms in a supervised way is data annotation. In case of image classification, datasets should be labeled as to what images constitute a positive image, i.e. an image containing solar panels, and what images constitute a negative image, i.e. an image without solar panels. In the case of object annotation, each image should be labeled with either bounding boxes or bounding polygons for each of the solar panels in the pictures. To train a deep learning algorithm a lot of manual labeled data is necessary. To simplify data annotation, on the one hand a tool for image classification was developed within this project. On the other hand, an off-the-shelf solution was used for the annotation for object detection. We will now shortly discuss both solutions.

\subsection{Annotation for Image Classification}

\begin{figure}[h]
    \centering
    \includegraphics[width=0.9\textwidth]{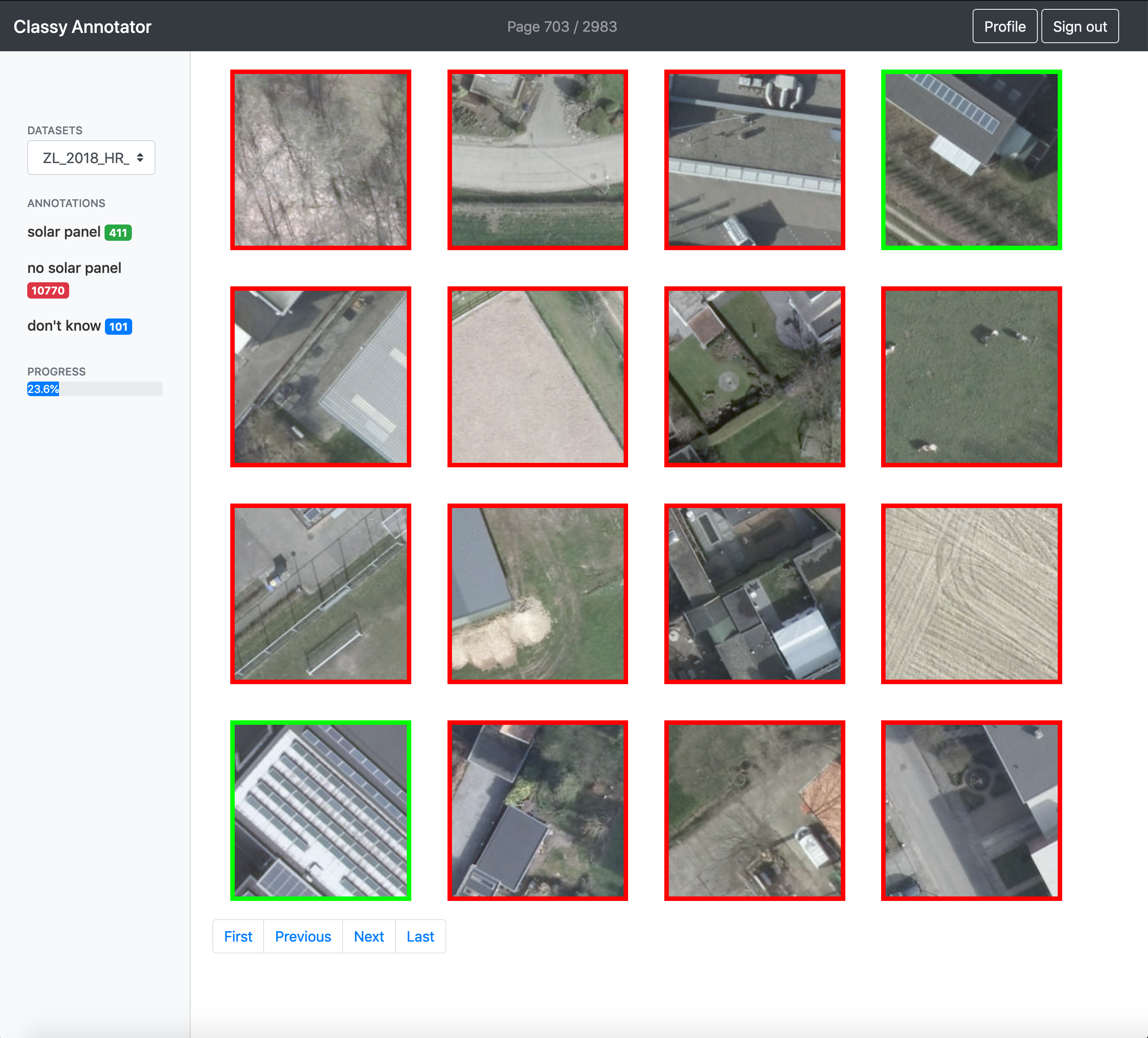}
    \caption{Screen shot of the tool developed for image classification. Users are presented with a grid of sixteen photos to annotate into three categories: solar panels (green), no solar panels (red), and do not know (blue)}
    \label{fig:tool_for_image_classification}
\end{figure}

Figure \ref{fig:tool_for_image_classification} shows a screen shot of the web-based tool we developed to annotate images for image classification. Images can be classified into three classes: solar panels (green color), no solar panels (red color), and do not know (blue color). By clicking an image, users can cycle between the three classes. Using a combo box the user can select a dataset to annotate, out of the datasets that were assigned to this user. Each dataset is presented in pages of sixteen images each. Each of these images is already pre-annotated into a certain class. This allows users to quickly scan the images and only change the annotations on images that have been incorrectly labeled. While all images could be manually pre-annotated with a certain class, in our case we used a neural network already trained to recognize solar panels to pre-annotate the images. Using a neural network with an already reasonable accuracy can speed up the annotation task greatly. This is especially important in large datasets that can be cumbersome to completely annotate manually.

Two design choices were taken to ensure annotation quality. On the one hand, a separate class for images that users are uncertain about was added. By adding a \emph{do not know} class, images with high uncertainty can be kept out of the dataset, and the dataset used for training and evaluation can be kept as clean as possible. On the other hand, by allowing each image to be annotated by multiple users, the uncertainty of the annotations can be measured. If user annotations are highly similar, the annotation has a high certainty. Conversely, if user annotations are dissimilar, the annotations is highly uncertain and should either be checked by an expert or discarded from the dataset.

\subsection{Annotation for Object Detection}

For the purpose of applying object detection algorithms, we used the VIA annotation tool \citep{dutta2019vgg} to annotate polygons in the 330x330 pixels images of NRW. A total of 5432 images were annotated by taking care of collecting the polygons of all the solar panels collected in the images and the associated bounding boxes. The annotation used by the object detection algorithm then contains $x,y,w,h$ tuples, where $x$, $y$ represent the coordinates of the center, and $w$, $h$ represent the width and height of the bounding box. 
Multiple bounding boxes per image are possible and they occur frequently, but this does not constitute a problem for object detection algorithms. The Mask R-CNN algorithm used in this report does not need to report the negative images, the algorithm learns to associate a confidence to the images and consequently, in a trained model, those bounding boxes that do not containing solar panels would have a low confidence score.

\begin{figure}
    \centering
    \includegraphics[width=0.3\textwidth]{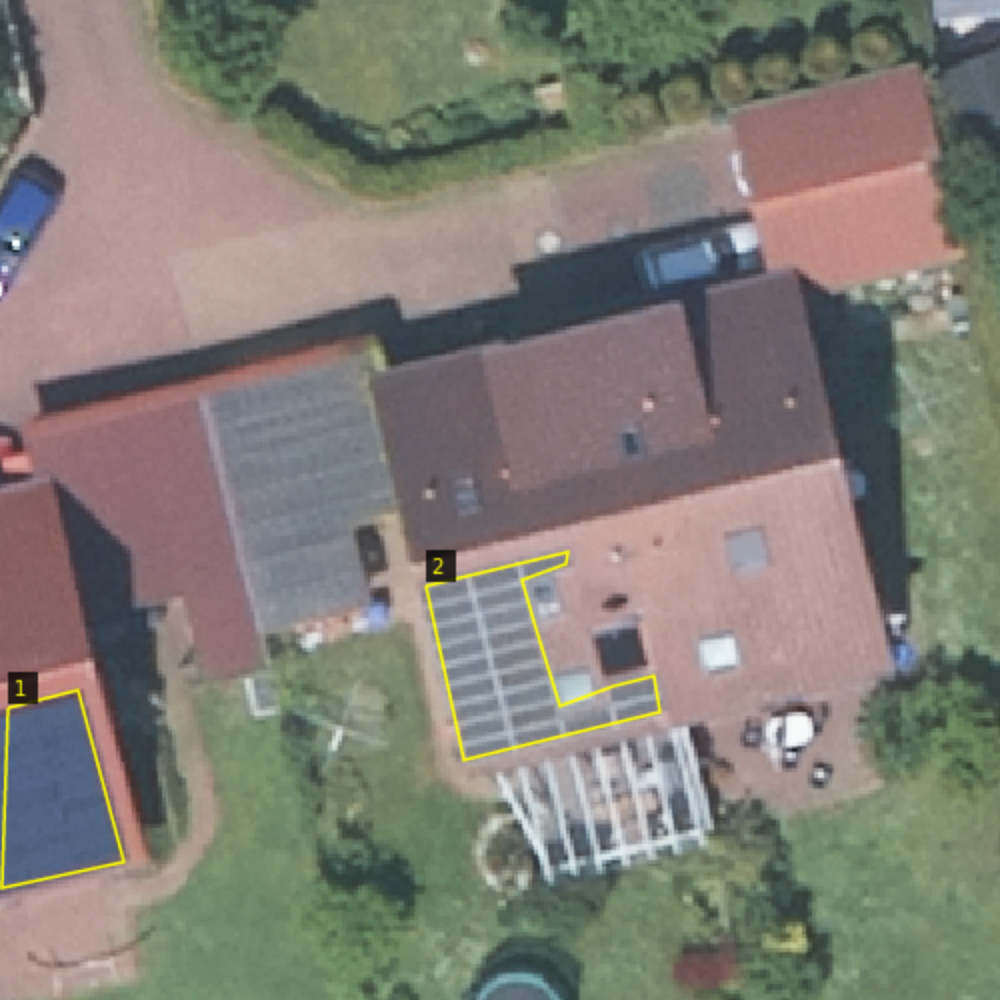}
    \includegraphics[width=0.3\textwidth]{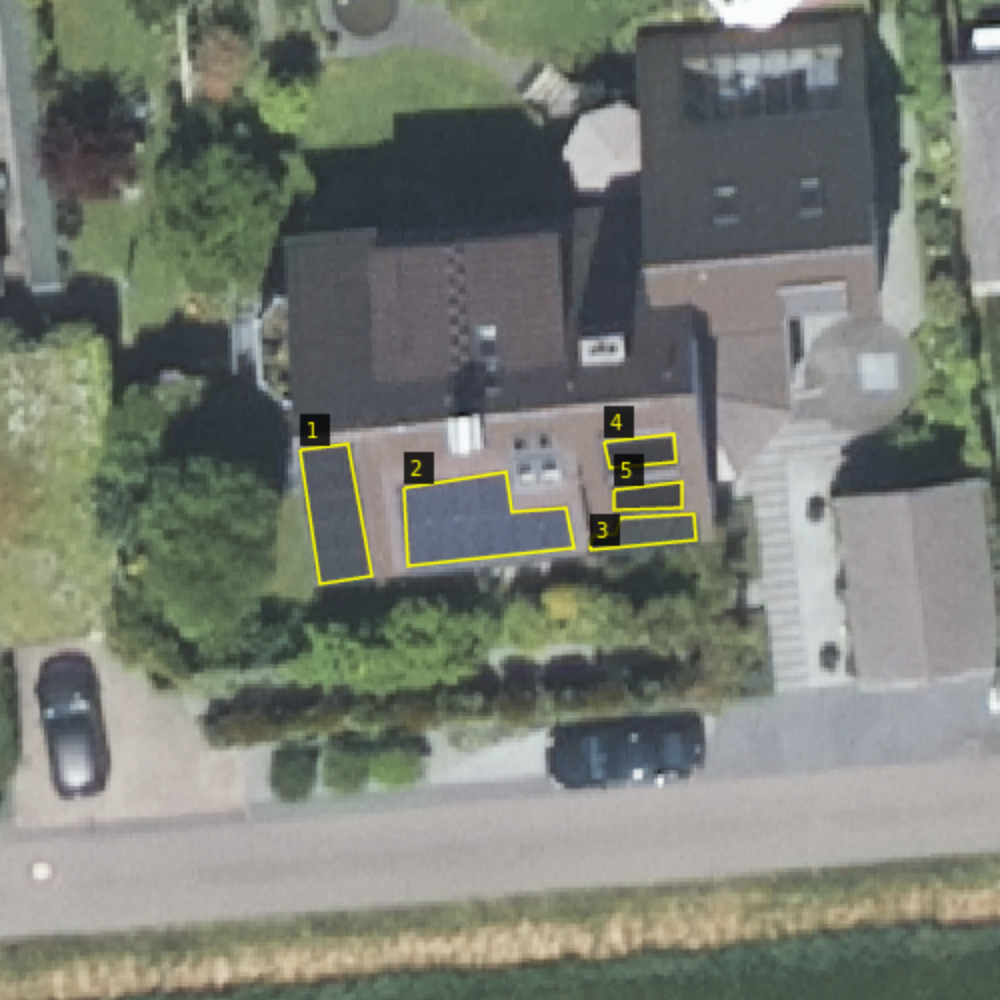}
    \includegraphics[width=0.3\textwidth]{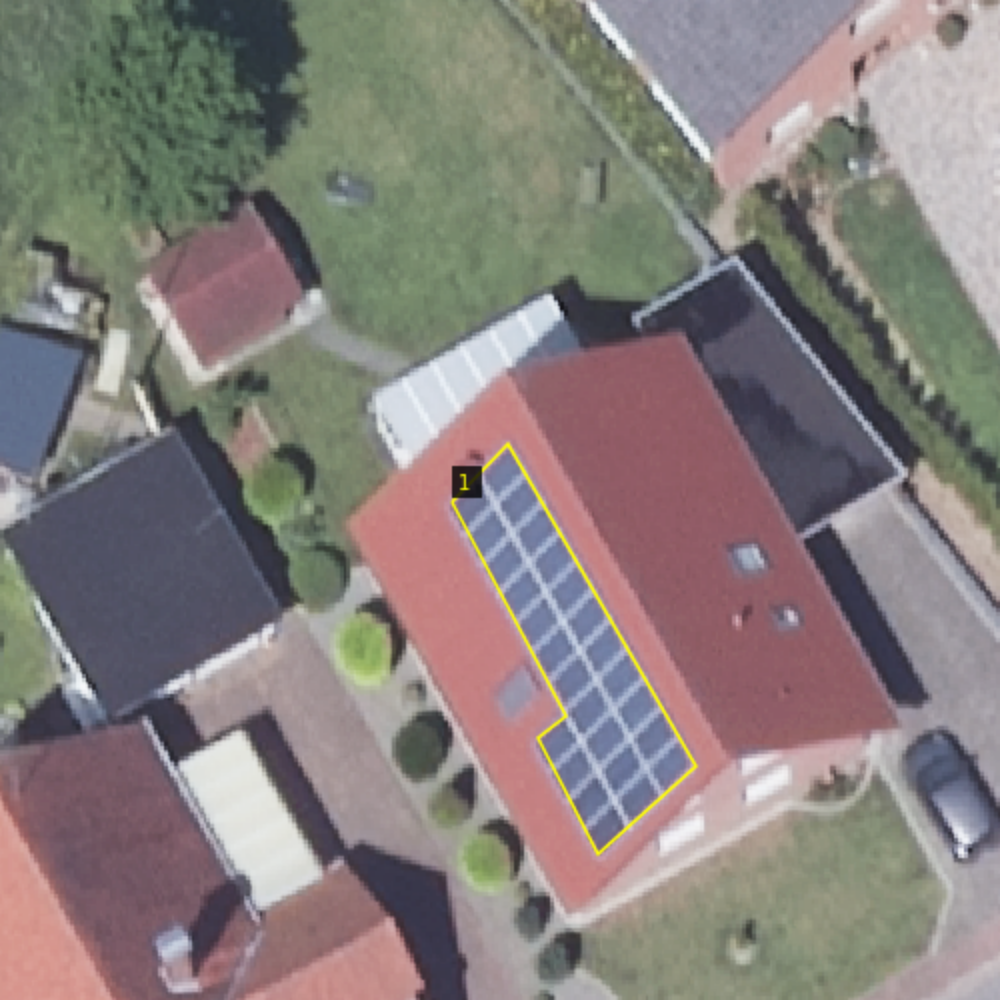}
    
    \caption{Object Detection Annotation Procedure}
    \label{fig:annot_object_detection}
\end{figure}

The object detection images have been annotated by with single experts sharing the load in a crowd sourcing environment. It was not reputed necessary to double annotate the images because these have been taken directly from the NRW registry where the location of the solar panels is known. This in itself limited the probability of including segmentations of structures that are not solar panels, and given that the shape of the solar panels is most of the time regular, once known that a solar panel is present in an area the annotation task did not present a challenge as in the case of segmenting medical images or material science images, where most of the time the object has a complex border.

\section{Model training}
\label{section:model_training}

As was mentioned in section \ref{section:deep_learning_algorithms} we trained two types of models in the DeepSolaris project: (1) models for image classification, and (2) models for object detection. Furthermore, we evaluated multiple model architectures on two distinct geographical regions. In this section, we will shortly outline the training procedure for both types of models and for both geographical regions.

\subsection{Model Description for Image Classification}
As mentioned in section~\ref{sec:TL}, existing pre-trained models are investigated and further trained on the aerial image datasets. Using these pre-trained models, four different scenarios were evaluated, two training the convolutional neural network on the Heerlen dataset and two training the convolutional neural network on the NRW dataset. 

\begin{itemize}
    \item using a convolutional neural network as a feature extractor on the Heerlen data
    \item fine tuning a convolutional neural network on the Heerlen data and evaluating it on the NRW data
    \item fine tuning a convolutional neural network on the NRW data and evaluating it on the Heerlen data
    \item training an auto-encoder architecture on the NRW data
\end{itemize}

\subsubsection{Training Setup for the Heerlen data}
A pre-trained VGG16 model was trained as a classifier on the Heerlen data. On the one hand, the VGG16 model was used as-is as a feature extractor. In this case, the neural network activations are taken at a specific layer and converted to a feature vector. In turn, this feature vector is used to train a machine learning classifier, like for example a logistic regression or a support vector machine (SVM). Using the network as-is gives an important base-line classifier performance that indicates how well a network trained on ILSVRC \citep{imagenet_cvpr09} will perform on another problem domain: in this case, solar panel classification. 

On the other hand, a pre-trained VGG16 was fine-tuned using the Heerlen data. In this case, the convolutional layers of the VGG16 model were taken and a classification layer (sigmoid/softmax) was added on top. After that, the network was fine-tuned by allowing it to train for a couple more epochs. 

\subsubsection{Training setup for the NRW data}
Two models were evaluated on the NRW data: InceptionResNetV2 and VGG16. In addition, we also tried an auto-encoder architecture. Except for the ad hoc auto-encoder, these architectures have been created during the last three years, are available in the Keras library with pre-trained weights from the ILSVRC \citep{imagenet_cvpr09}, and together they present the most important recent improvements in CNNs, as inception modules and residual layers. Still there are substantial differences in their network’s architectures, as for example their number of parameters and layers (see table 1 below). For more detailed information, see \citep{curier2018monitoring}. 

In this study, InceptionResNetV2, and VGG16 models with weights from the ILSVRC \citep{imagenet_cvpr09} as baseline will be trained on our new datasets. The pre-trained models were adapted for the specific case of solar panel detection, the output layers of the models are extended by addition of a global average pooling to reduce the dimension, a fully connected layer with 512 nodes, and finally a sigmoid layer was appended.

\subsection{Model Description for Object Detection}

Within this project, we made use of the Mask R-CNN algorithm \cite{he2017mask} with a pre-trained model (on MS COCO database \cite{lin2014microsoft}) by fine tuning it on the images annotated with the bounding box of the solar panels. With respect to the simplest case with one single bounding box R-CNN also provides a proposition of a number of regions of interest in which the objects may be situated. 

\subsubsection{Training setup for Object Detection}

From the perspective of Object Detection, we also make use of transfer learning by retraining the Mask R-CNN network available from a Github repository \footnote{\url{https://github.com/matterport/Mask_RCNN}}. The Mask R-CNN model has been trained on the available images for 5 epochs.

\subsection{Evaluation}
To evaluated the models during training and validation several standard metrics were used: the precision, the recall and F1 score. The \emph{precision} expresses the percent of the correctly classified images from the actual solar panel images i.e. how many of those instances identified as positives are actually positive. The \emph{recall} states the percent of the actual solar panels are recognized as such i.e. how many of the positive instances in the population were identified correctly by the model. The \emph{F1} score is the harmonic average of the precision and recall, where an F1 score reaches its best value at 1 (perfect precision and recall) and worst at 0. In case of binary classification, the \emph{F1-score} is a measure of a test's accuracy and is useful when you want to seek a balance between precision and recall. In this report, we provided the precision, recall and F1-score independently for both the positive and the negative case. To calculate the metrics the scikit-learn module's \texttt{classification\_report} function was used; along with all the metrics it also provides an overview of the total number of samples, the samples per class, the micro average, the macro average, and the weighted average. The micro average averages the total true positives, false negatives and false positives. In the scikit-learn module the micro average for a binary problems has the same formula as accuracy, therefore we replaced the micro average by accuracy in all tables in the result section. The macro average averages the unweighted mean per label, whereas the weighted average averages the mean weighted by the support, or the number of samples per class, per label. Finally, during model training, loss and accuracy curves were compared for both the training set and the validation set. This allowed us to spot overfitting and underfitting already during the training process.

\section{Validation}
\label{section:validation}

During model training, model performance was evaluated on images that were taken from the same geographical area. To create official statistics using deep learning models, it is important that model performance also scales well to other geographical regions; we would like a model that was trained in one area of a country to perform similarly in another. To evaluate how well model performance generalizes to other geographical regions, several validation steps can be carried out:
\begin{itemize}
    \item validate the model performance on the \emph{same aerial image} but on a distinct geographical area,
    \item validate the model performance on a \emph{different aerial image}, like for example a different country or a different year,
    \item comparing the model predictions to the available register data.
\end{itemize}
We will detail each of these validation steps in the subsections below.

\subsection{Validation on the same aerial image}
By evaluating model performance on the same aerial image, but on other geographical areas and images that have not been seen before, we validate how well the model is suited to deal with the changes within a country. We test if the model is resilient to differences in geography, building style, and urbanization. By using the same aerial image, the effects of different image color distributions caused by normalization and camera equipment can be minimized. It follows, that the effects of differences in geographical patterns can be thus be better analyzed by using the same image data source.  

\subsection{Cross validation on a different aerial image}
By evaluating the model on different aerial images, we can validate how well the model fares in situations that are really distinct. When using an aerial image from the \emph{same geographical region} but from a different year, the resilience of the model against year-by-year changes in the aerial picture can be tested. We can measure how well the model can deal with different image color distributions caused by differences in normalization and differences in camera equipment. Conversely, when using an aerial image from a \emph{different county}, next to measuring the resilience to different image color distributions, we furthermore test how well the model deals with changes in countries' building style, architecture, and urban planning.

\subsection{Comparisons to registers}
Finally, by comparing the model predictions to the register data that was collected by our NSOs, we can validate the quality of both the model and the register. For the situations in which the model prediction and register are consistent, we can assume that the results are correct. However, by analyzing the situations were the model prediction and the register are different, we can either improve model performance when its prediction was incorrect, or correct the register if an address was found that was incorrectly labeled in the register.

\section{Related Work}
\label{section:recent_work}

Remote sensing provides a way to observe objects and geographical areas from afar, without making physical contact with the observed. In general, remote sensing entails observation from either satellite or aircraft-based sensors. In this sense, a distinction can be made between active and passive remote sensing. Active remote sensing emits a signal and measures the reflected signal. In contrast, passive remote sensing observes radiation that is reflected or emitted by the object itself. The aerial images we will use in this paper are an example of passive remote sensing; the sunlight reflected by the earth is registered by the camera of the airplane creating the aerial images. Throughout the years, remote sensing has been used for a range of applications, some of which will shortly discuss now. 

First, remote sensing is used for the observation of \emph{natural phenomena}. \citep{DBLP:journals/corr/LiuRPCKLKWC16} presents a deep-learning algorithm to help detect extreme weather events in climate observation. Several types of polution can also be detected from remote sensing sources. \citep{brekke2005oil} for example, gives an overview of all kinds of techniques that can be used to detect oils spills. Moreover, satellite imagery can be analysed with convolutional neural networks to detect forest boundaries, forest density, and deforestation \citep{rs11070768, ortega2019evaluation, Caballero_Espejo_2018}. \citep{chi2017} presents a deep learning approach to predict the Artic sea ice concentration. Second, remote sensing also provides an overview of \emph{natural resource management}, more specific land use and land cover. An example of a deep learning algorithm trained for cloud, shadow, and land cover scenes that was evaluated for geographic independence and also across datasets is given in \citep{SHENDRYK2019124}.  Related to that, \citep{DBLP:journals/corr/abs-1709-00029} provides a dataset and benchmark for convolutional neural networks for land use and land cover classification. In addition, crop health, status, and yield are often studied with remote sensing data \citep{rs11232769, rs10122007, rs10111726}. Third, remote sensing can be used to explore \emph{historical settlements} that may have been obscured and are not visible from the ground level. Satellite imagery and LiDAR data can be used to unveil undiscovered parts of Mayan settlements under the jungle canopy \citep{weishampel2010remote, chase2010lasers}. Likewise, Roman settlements were also analysed and/or rediscovered using a combination of remote sensing techniques \citep{mozzi2016roman, Garcia_Garcia_2017}. Last, remote sensing can also be used to explore \emph{current day settlement and urban development}. Building footprint can be extracted from satellite imagery with machine learning techniques \citep{rs11232803}. Furthermore, urban characteristics and development can also be followed using satellite imagery \citep{MATHIEU2007179, haverkamp2002extracting, malmir2015urban}. Another example is the mapping of wealth/poverty on the basis of remote sensing information \citep{Georganos_2019, Zhao_2019, Wang_2019}. All in all, remote sensing data provides accurate and timely information about natural, urban and societal developments and as such are an interesting source of information for official statistics. 

One specific example of an application of remote sensing for official statistics is generating statistics about renewable energy. Several studies have already described approaches to automate the detection of photo-voltaic installations. For example, an automated way to detect photo-voltaic installations in satellite images of a specific area of California is presented in \citep{malof2015}. They extract a list of hand-crafted features from the image and train an SVM classifier on top of these features to identify pixels that belong to solar panels and combine them into regions. Likewise, \citep{Malof2016} presents another approach that trains a Random Forest pixel-based classifier for aerial images on top of hand-crafted features. Another model based on hand-crafted features is presented in \citep{Puttemans2016} where several pixel-based approaches for the detection of photo-voltaic installations are compared. More recently, research has focused more on deep-learning based approaches as in recent years they have shown to greatly outperform the approaches based on hand-crafted features. For example, \citep{Malof2017, DBLP:journals/corr/abs-1801-04018} train deep-learning segmentation models on high resolution aerial images of certain parts of California. Similarly, \citep{Yu2018} trains a Inception V3   \citep{DBLP:journals/corr/SzegedyVISW15} to first classify images for the presence of photo-voltaic panels across the entire United States. For the images that do contain photo-voltaic panels, a segmentation is generated by using class activation maps \citep{zhou2015cnnlocalization, DBLP:journals/corr/SelvarajuDVCPB16} generated from the feature maps of the network. While each of these publications shows promising results, the validation of the model performance is either limited to a small geographical area \citep{malof2015, Malof2016, DBLP:journals/corr/abs-1801-04018, Malof2017, Puttemans2016}, or is focused at one country (however large) \citep{Yu2018} with similar building styles and urban planning regulations. An especially interesting paper in this aspect is \citep{Wang2017}, in which the authors compare CNN performance when trained and tested on the same city vs. training on one city and testing on another. In this report, we aim to take this approach a step further by first training and testing the algorithms on different (nearby) geographical regions in the same country and then evaluating the trained algorithms in a cross-border fashion. More specific, we will present a cross-border evaluation of a deep-learning model for both classification and object-detection of photo-voltaic installations in two countries in the European Union: the Netherlands and Germany.

\chapter{Results}
\label{chapter:results}
\label{sec:results}

This chapter presents findings concerning the use of deep learning architectures on aerial images for the area of NRW and Limburg. The evaluation was carried out in five phases. In phase I, we tried out several state-of-the-art algorithms on DOP25 aerial images (see \ref{section:results_phase1}). In phase II, we continued with using a VGG16 network as a feature extractor on the DOP10 images of Heerlen (see section \ref{section:fixed_feature_extractor}) to measure how well VGG16 performs out-of-the-box on the task of solar panel classification. In phase III, we trained several classifiers on both North Rhine Westphalia and Limburg. Section \ref{section:fine_tuning_nrw} describes training an image classification model, as well as an object detection model on North Rhine Westphalia. Section \ref{section:fine_tuning_vgg16} describes fine-tuning a VGG16 model on Heerlen. After that, we cross validate the models by evaluating the models trained on Heerlen on North Rhine Westphalia, and vice versa (see section \ref{section:phase4_blind_test}). Last, we present our evaluation of the use of satellite data for the detection of solar panels in section \ref{section:phase5_satellite_images}.

\section{Phase I: First Approach}
\label{section:results_phase1}
\begin{figure}
    \centering
    \includegraphics{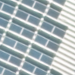}
      \includegraphics{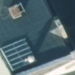}
    \caption{75x75 images used in phase 1.}
    \label{fig:phase1}
\end{figure}

In Phase I, we used ImageNet networks for the classification of images of NRW with a resolution of DOP20 and a size of $75\times75$ pixels. Despite the fact that the initial evaluation seemed to confirm that neural networks can in principle identify solar panels, we quickly came to the conclusion that the size of the images and the approach used to create validation and testing data sets was not reflecting the natural distribution of the solar panels in the images. For example if we take two positive images as shown in figure \ref{fig:phase1} the main issue with
these images is that the surface of the solar panel covers most of the image and it is often very close to the center.

Convolutional networks are not completely position or rotation invariant, so if the convolutional filters are fed with images that are homogeneous in terms of positions of the solar panels, the result is that the filters can only recognize solar panels if presented with the exact same situation. As a consequence, we quickly noticed that when we applied the trained convolutional models in the NRW data we obtained a degraded performance in terms of both false positives and false negatives. The consequence has therefore been that we decided to enlarge the resolution (to DOP10) and the size of the pictures (from 75 pixels to 200 pixels for Limburg and 330 pixels for NRW), in this way the occurrence of solar panels in the image would not be limited to the center of the image. In addition, we also decided to perform blind tests in which we apply the trained algorithms in never seen areas (see phase 4, phase 4b, and phase 4c of this report), in order to understand the difference between the real performance and the performance calculated on data. 

\section{Phase II: Using a pre-trained network as fixed feature extractor for the Heerlen dataset}

\label{section:fixed_feature_extractor}

In feature extraction, a convolutional neural network is used as is, with pre-trained weights on ImageNet, and without fine tuning the weights to a different problem domain. The features are extracted at a specific layer in the network: features can be extracted right after the convolutional layers but also after the fully connected layers. A one-dimensional feature vector is retrieved from the activations of the chosen layer and then a machine learning classifier is trained on top. Using a pre-trained convolutional neural network as a feature extractor gives a baseline out-of-the-box classification performance for the new problem domain: the pre-trained weights are used as-is without any re-training to the new problem domain. In the case of a solar panel classifier, using a pre-trained convolutional neural network as a feature extractor thus gives a measure of how well the convolutional filters trained on ImageNet are transferable to aerial pictures and the classification of solar panels. As such, several experiments were carried out using different neural network infrastructures as feature extractors and training a machine learning classifier on top. Using the baseline out-of-the-box performance the network infrastructures can be compared on their suitability for the classification of solar panels. Feature extraction was performed on every image in the dataset, after which the dataset was split using stratification in a training set of 75\% (or 10649 images) and a test set of 25\% (or 3551 images). After feature extraction a logistic regression as well as an SVM classifier were trained and evaluated. Several experiments were carried out of which the most successful are reported here.

First, a pre-trained VGG16 was used as a feature extractor. In the first experiments, the features were extracted right after the maxpooling layer of the last convolutional block of VGG16 (block 5). In the subsequent experiments, the features were extracted after block4. The sizes of the activation map between block4 and block5 differ and thus the sizes of the feature vectors vary too: the experiments were carried out to see whether a different size of feature vector would influence classification accuracy. 

\begin{table}[]
    \centering
    \begin{tabular}{c|c|c|c|c}
         & precision  & recall & f1-score &  support  \\
     \hline
     0   &    0.86    &  0.90   & 0.88 &      2495 \\
     1   &    0.73    & 0.67  & 0.70 &      1056 \\
     \hline
     accuracy &   &   & 0.83 &     3551 \\
     macro avg & 0.80 &  0.78  & 0.79 &     3551 \\
     weighted avg & 0.82 & 0.83 & 0.83 &  3551 
    \end{tabular}
    \caption{Classification metrics for feature extraction using all convolutional layers from a VGG16 and a Logistic Regression, best of trained models with C=100 }
    \label{tab:feature_vgg16_all_lr_results}
\end{table}

\begin{table}[]
    \centering
    \begin{tabular}{c|c|c|c|c}
          & precision &   recall &  f1-score &  support \\
          \hline
           0 & 0.86 & 0.89 & 0.87 & 2495 \\
           1 & 0.71 & 0.65 & 0.68 & 1056 \\
          \hline
   accuracy & & & 0.82 & 3551 \\
   macro avg & 0.79 & 0.77 & 0.78 & 3551 \\
weighted avg & 0.81 & 0.82 & 0.82 & 3551 \\
    \end{tabular}
    \caption{Classification metrics for feature extraction using all convolutional layers from a VGG16 and a Linear SVM, best of trained models with C = 0.1}
    \label{tab:feature_vgg16_all_svm_results}
\end{table}{}

\begin{table}[]
    \centering
    \begin{tabular}{c|c|c|c|c}
         & precision  & recall & f1-score &  support  \\
     \hline
     0   &    0.86    &  0.89  & 0.87 & 2495 \\
     1   &    0.72    &  0.65  & 0.68 & 1056 \\
     \hline
     accuracy & &  & 0.82 & 3551 \\
     macro avg & 0.79 &  0.77  & 0.78 & 3551 \\
     weighted avg & 0.82 & 0.82 & 0.82 &  3551 
    \end{tabular}
    \caption{Classification metrics for feature extraction using the convolutional layers until block4\_pool from a VGG16 and a Logistic Regression, best of trained models with C=10}
    \label{tab:feature_vgg16_block4_lr_results}
\end{table} 

\begin{table}[]
    \centering
    \begin{tabular}{c|c|c|c|c}
         & precision & recall & f1-score & support \\
       \hline     
       0 & 0.86 & 0.89 & 0.87 & 2495 \\
       1 & 0.72 & 0.65 & 0.68 & 1056 \\
      \hline
   accuracy & & & 0.82 & 3551 \\
   macro avg & 0.79 & 0.77 & 0.78 & 3551 \\
weighted avg & 0.82 & 0.82 & 0.82 & 3551 \\
    \end{tabular}
    \caption{Classification metrics for feature extraction using the convolutional layers until block4\_pool from a VGG16 and a SVM, best of trained models with C=0.1}
    \label{tab:feature_vgg16_block4_svm_results}
\end{table}

Table \ref{tab:feature_vgg16_all_lr_results} shows the results of training a logistic regression classifier on the features extracted after VGG16 block5, whereas table \ref{tab:feature_vgg16_all_svm_results} shows the same results for a SVM classifier. As can be seen, the logistic regression classifier slightly outperforms the SVM. Conversely, from table \ref{tab:feature_vgg16_block4_lr_results} and table \ref{tab:feature_vgg16_block4_svm_results} for the feature extraction after block 4, the performance of the logistic regression is exactly the same as the performance using SVM. In addition, the results do not differ drastically (if at all) between the feature extraction at block4 versus the feature extraction at block5. The logistic regression trained with the features extracted from block5 performs best, but differences between evaluation metrics is at most 2\%. In all cases it can be seen that the classification of the negative cases is more successful than the classification of the positive cases. All in all, out-of-the-box classification for the negative cases performs already quite well. However, performance for the positive case tops at around 73\% precision for the logistic regression with features from block5, and still leaves much room for improvement.

Second, several combinations of Xception were evaluated. Again, the classification performance was evaluated when extracting features at different layers in the network. On the one hand, classification performance was evaluated when Xception was used as a feature extractor until the global average pooling layer. On the other hand, classification performance was evaluated when features were extracted at the last fully connected layer in front of the global average pooling layer. The results when using logistic regression with the features from the global average pooling in table \ref{tab:xception_lr_gap} outperform the logistic regression with features from the last layer (see table \ref{tab:xception_lr_last_layer} in terms of precision for the negative case and f1-score for the positive. However, for the negative/positive recall and negative f1-score the logistic regression with features from the last layers perform better. Both logistic regression classifiers have the same performance for positive precision. For the SVM classifier, the differences are more pronounced with the SVM trained on the last layer before the global average pooling having in general a better performance. The SVM with global average pool features performs better on negative precision and positive recall (see table \ref{tab:xception_svm_gap}) while the SVM with last layer before features performs better on negative recall, positive precision, and negative and positive f1-score (see table \ref{tab:xception_svm_last_layer}). These results are also reflected in the average precision, recall, and f1-score which do not provide a clear separation between layer location or classifier type either.

\begin{table}[]
    \centering
    \begin{tabular}{c|c|c|c|c}
          & precision & recall & f1-score & support \\
       \hline
       0  & 0.86 & 0.69 & 0.77 & 2495 \\
       1  & 0.50 & 0.73 & 0.59 & 1056 \\
       \hline
   accuracy & & & 0.70 & 3551 \\
   macro avg & 0.68 & 0.71 & 0.68 & 3551 \\
weighted avg & 0.75 & 0.70 & 0.71 & 3551 \\
    \end{tabular}
    \caption{Classification metrics for Xception Global Average Pooling Logistic Regression, C = 10000.0}
    \label{tab:xception_lr_gap}
\end{table}{}

\begin{table}[]
    \centering
    \begin{tabular}{c|c|c|c|c}
             & precision & recall & f1-score & support \\
           \hline     
           0 & 0.89 & 0.39 & 0.54 & 2495 \\
           1 & 0.38 & 0.89 & 0.53 & 1056 \\
           \hline
   accuracy & & & 0.54 & 3551 \\
   macro avg & 0.64 & 0.64 & 0.54 & 3551 \\
weighted avg & 0.74 & 0.54 & 0.54 & 3551 \\
    \end{tabular}
    \caption{Classification metrics for Xception Global Average Pooling SVM
, C = 10000.0}
    \label{tab:xception_svm_gap}
\end{table}

\begin{table}[]
    \centering
    \begin{tabular}{c|c|c|c|c}
             & precision & recall & f1-score & support \\
           \hline
           0 & 0.81  &  0.76 & 0.78 & 2495 \\
           1 & 0.50 & 0.58 & 0.54 & 1056 \\
          \hline
   accuracy & & & 0.71 & 3551 \\
   macro avg & 0.66 & 0.67 & 0.66 & 3551 \\
weighted avg & 0.72 & 0.71 & 0.71 & 3551 \\
    \end{tabular}
    \caption{Classification metrics for Xception Last Layer before GAP, Logistic Regression, C = 0.1}
    \label{tab:xception_lr_last_layer}
\end{table}{}

\begin{table}[]
    \centering
    \begin{tabular}{c|c|c|c|c}
                    & precision & recall & f1-score & support \\
           \hline
           0 & 0.78 & 0.77 & 0.78 & 2495 \\
           1 & 0.47 & 0.49 & 0.48 & 1056 \\
           \hline
   accuracy & & & 0.69 & 3551 \\
   macro avg & 0.63 & 0.63 & 0.63 & 3551 \\
weighted avg & 0.69 & 0.69 & 0.69 & 3551 \\
    \end{tabular}
    \caption{Classification metrics for Xception Last Layer before GAP SVM, C = 100.0}
    \label{tab:xception_svm_last_layer}
\end{table}

In general, when using Xception as a feature extractor the classification of the negative cases performs a lot better than the classification of the positive cases. It can also be seen that Xception performs (a lot) poorer as a feature extractor than VGG16. It is hard to pinpoint a clear reason for VGG16 drastically outperforming Xception. First, it might be the case that the features present in Xception are less suitable for the domain of solar panel classification than those present in VGG16. Second, the features might have been extracted at the wrong layers; because Xception is a non-linear network it is much harder to predict where to extract the features. Last, the drastic differences in precision, recall, and f1-score between the negative and positive cases suggest that the features trained on ImageNet are not discriminative enough for the domain of solar panel classification. However, all experiments provide an overview of the base line out-of-the-box performance to expect when these networks are trained using fine-tuning and different fully connected layers. The results of fine-tuning a pre-trained VGG16 network will be presented in section \ref{section:fine_tuning_vgg16}.

\section{Phase IIIa: Transfer Learning on NRW Data with Inception2 and custom architectures}

\label{section:fine_tuning_nrw}

This section presents the first findings for the application of five CNN models: VGG16, InceptionV2, and an auto-encoder model. These models are publicly available in the Keras library. The models are trained on aerial photographs of different urban landscapes, using existing registers to train and validate the developed algorithms.

The starting point for the models is the use of weights from the ImageNet Large Scale Visual Recognition Competition (ILSVRC) \citep{imagenet_cvpr09}. In this competition, different algorithms are tested on how well they perform in the detection of objects and the classification of different images. Detection of solar panels is not part of this competition. Therefore VGG16 and InceptionResNet2 will use pre-trained weights and substitute the last layer of the network.

For the purposes of the testing, we used the LANUV registry to identify positive solar panels and train our algorithms, we could identify 50 000 images with solar panels. We therefore collected a sample of 10 000 positive images from LANUV and 10 000 negative images from the NRW WMS services, then we took a test set of about a 2000 images still balanced, plus a validation set of 1000 images which instead is unbalanced in order to understand the ability of the networks to detect solar panels and to discriminate those from negative images.

\subsection{Experiment 1: Evaluation on Data, NRW}
In this experiment, we investigate to what extent models that have already been trained for other applications can also be used for the detection of solar panels. To this end, the VGG16 and InceptionResNet2 for different options of transfer learning are considered. The results are presented in below. We also show how a Mask RCNN algorithm performs when we use its confidence about predicted solar panels to take decisions on whether an image contains or not a solar panel.
To account for the fact that the second data set is imbalanced, we also consider the accuracy, macro and weighted averages 
(Notice that since version 0.22 of sklearn, used in this report, the micro-avg is reported as accuracy when the classification is binary, as the formula is the same for the binary case).
In a balanced data set, such averages would all be the same or close to each other (as in table \ref{table:vgg}), but in an imbalanced data set these may vary considerably (as in table \ref{table:vggval}), providing an understanding of the behaviour of the algorithm in natural settings.

\begin{table}[h!]
    \centering
    \begin{tabular}{c|c|c|c|c}
         & precision & recall & f1-score & support \\
         \hline
           0 &  0.87 & 0.63 & 0.73 & 1040 \\
           1 &  0.71 & 0.91 & 0.80 & 1062 \\
        \hline
  accuracy &  &  & 0.77 & 2102 \\
   macro avg & 0.79 & 0.77 & 0.76 & 2102 \\
weighted avg & 0.79 & 0.77 & 0.76 & 2102 \\
    \end{tabular}
    \caption{Classification metrics for a fine-tuned VGG16 network for the NRW test set}
    \label{table:vgg}
\end{table}

\begin{table}[h!]
    \centering
    \begin{tabular}{c|c|c|c|c}
         & precision & recall & f1-score & support \\
         \hline
           0 &  0.61 & \textcolor{red}{0.29} & 0.39 & 189 \\
           1 &  0.86 & 0.96 & 0.91 & 876 \\
        \hline
   accuracy & &  & 0.84 & 1065 \\
   macro avg & 0.74 & 0.62 & 0.65 & 1065 \\
weighted avg & 0.82 & 0.84 & 0.82 & 1065 \\
    \end{tabular}
    \caption{Classification metrics for a fine-tuned VGG16 network for the NRW validation set}
    \label{table:vggval}
\end{table}

The first attempted network is the VGG16 one (see table \ref{table:vgg}). By using a balanced data set, we obtain a balanced precision and recall of 77\%, when we set the prediction confidence threshold to 0.5. When trying the model on a further validation data set, we find out that the network does not generalize well to new data as shown in table \ref{table:vggval}. In particular, the validation data set has been taken with a prominence of solar panels in order to understand how well the network responds to the positive class. The main issue is that albeit the model can recognize images with solar panels, it struggles to recognize negatives.

\begin{table}[h!]
    \centering
    \begin{tabular}{c|c|c|c|c}
         & precision & recall & f1-score & support \\
         \hline
           0 &  0.88 & 0.93 & 0.90 & 1040 \\
           1 &  0.92 & 0.88 & 0.90 & 1062 \\
        \hline
   accuracy &  &  & 0.90 & 2102 \\
   macro avg & 0.90 & 0.90 & 0.90 & 2102 \\
weighted avg & 0.90 & 0.90 & 0.90 & 2102 \\
    \end{tabular}
    \caption{Classification metrics for a fine-tuned InceptionResNetV2 network for the NRW test set}
    \label{tab:inc}
\end{table}

\begin{table}[h!]
    \centering
    \begin{tabular}{c|c|c|c|c}
         & precision & recall & f1-score & support \\
         \hline
           0 &  0.61 & \textcolor{red}{0.40} & 0.48 & 189 \\
           1 &  0.88 & 0.94 & 0.91 & 876 \\
        \hline
   accuracy &  &  & 0.85 & 1065 \\
   macro avg & 0.74 & 0.67 & 0.70 & 1065 \\
weighted avg & 0.83 & 0.85 & 0.84 & 1065 \\
    \end{tabular}
    \caption{Classification metrics for a fine-tuned InceptionResNetV2 network for the NRW validation set}
    \label{tab:incval}
\end{table}

As shown in Table \ref{tab:inc} and Table \ref{tab:incval} a better result can be obtained when using the InceptionResNet2 architecture. The same problem with the false positives presents itself also for this model, although the results happen to be better than with the VGG16 network.

If we try with auto-encoders, due to the fact that they cannot be trained in a supervised manner, they actually present a much worse ability to generalize, as shown in tables \ref{tab:autenv} and \ref{tab:autenval}. Given these results, we opted for the InceptionResNet2 for the blind test on the NRW and Limburg data.

\begin{table}[h!]
    \centering
    \begin{tabular}{c|c|c|c|c}
         & precision & recall & f1-score & support \\
         \hline
           0 &  0.88 & 0.47 & 0.61 & 1040 \\
           1 &  0.64 & 0.94 & 0.76 & 1062 \\
        \hline
   accuracy &  &  & 0.71 & 2102 \\
   macro avg & 0.76 & 0.70 & 0.69 & 2102 \\
weighted avg & 0.76 & 0.71 & 0.69 & 2102 \\
    \end{tabular}
    \caption{Classification metrics for an auto encoder network for the NRW test set}
    \label{tab:autenv}
\end{table}

\begin{table}[h!]
    \centering
    \begin{tabular}{c|c|c|c|c}
         & precision & recall & f1-score & support \\
         \hline
           0 &  0.37 & 0.10 & 0.15 & 189 \\
           1 &  0.83 & 0.96 & 0.89 & 876 \\
        \hline
   accuracy &  &  & 0.81 & 1065 \\
   macro avg & 0.60 & 0.53 & 0.52 & 1065 \\
weighted avg & 0.75 & 0.81 & 0.76 & 1065 \\
    \end{tabular}
    \caption{Classification metrics for an auto encoder  network for the NRW validation set}
    \label{tab:autenval}
\end{table}

If instead we use object detection, by taking as a condition for classifying an image as having at least a detected solar panel region with a score with 90\% confidence as a threshold, we obtain table \ref{tab:maskrnnbal} and for the unbalanced data set we obtain table \ref{tab:maskrnnunbal}.

\begin{table}[h!]
    \centering
    \begin{tabular}{c|c|c|c|c}
         & precision & recall & f1-score & support \\
         \hline
           0 &  0.88 & 0.93 & 0.90 & 1040 \\
           1 &  0.92 & 0.88 & 0.90 & 1062 \\
        \hline
  accuracy &  &  & 0.90 & 2102 \\
   macro avg & 0.90 & 0.90 & 0.90 & 2102 \\
weighted avg & 0.90 & 0.90 & 0.90 & 2102 \\
    \end{tabular}
    \caption{Classification metrics for a Mask RCNN object detection model on a balanced data set.}
    \label{tab:maskrnnbal}
\end{table}

\begin{table}[h!]
    \centering
    \begin{tabular}{c|c|c|c|c}
         & precision & recall & f1-score & support \\
         \hline
           0 &  0.71 & \textcolor{red}{0.89} & 0.79 & 189 \\
           1 &  0.98 & 0.92 & 0.95 & 876 \\
        \hline
  accuracy &  &  & 0.92 & 1065 \\
   macro avg & 0.84 & 0.91 & 0.87 & 1065 \\
weighted avg & 0.93 & 0.92 & 0.92 & 1065 \\
    \end{tabular}
    \caption{Classification metrics for a Mask RCNN object detection model on a imbalanced data set.}
    \label{tab:maskrnnunbal}
\end{table}

It would seem that InceptionV2 and Mask RCNN are the best candidates to detect images that may contain new solar panels, in the following we therefore evaluate these two algorithms on NRW data which is then controlled by human experts.

\section{Phase IIIb: Using a pre-trained VGG16 for fine-tuning}

\label{section:fine_tuning_vgg16}

As seen in section \ref{section:fixed_feature_extractor} using pre-trained convolutional neural networks as a feature extractor left much to be desired. Especially, classification performance for the positive case topped at 73\%. In this section, a convolutional neural network will be fine-tuned on the Heerlen dataset in an attempt to increase classification performance. In specific, a VGG16-based network will be trained as it turned out to be the best performing feature extractor. To fine-tune the network, the dataset was split equivalently to section \ref{section:fixed_feature_extractor} into a training set (75\%) and a test set (25\%). After that, the convolutional layers of a VGG16 network, with weights pre-trained on ImageNet \citep{ILSVRC15}, were used as the basis for creating new convolutional neural networks structures. Different network structures were evaluated along with different forms of regularization and network configurations. All of the experiments started with finding the correct hyper-parameters for the model combined with the data set. The loss and accuracy curves were compared to spot if the model is overfitting/underfitting. Also, for each experiment, a confusion matrix was created and a classification report was generated with precision, recall, and F1 scores. The different experiments carried out will be shortly discussed below. 

\subsection{Fully connected layers}

Various neural network classifiers were evaluated on top of the VGG16 convolutional layers. As was the case for feature extraction in section \ref{section:fixed_feature_extractor}, the cut-off point of the original convolutional layers of VGG16 was varied. In the first case, the convolutional layers until the block4\_pool layer were used, in the second case the convolutional layers until the last block (block5\_pool) were used. On top of these convolutional layers, several combinations of fully connected layers were used, varying from:
\begin{itemize}
    \item adding one classification layer directly after the convolutional layers,
    \item adding a fully connected layer with varying sizes and a classification layer,  
    \item adding two fully connected layers with varying sizes and a classification layer.
\end{itemize}
For the classification layer both a sigmoid function and a softmax layer were evaluated.

\subsection{Regularization}
Different forms of regularization were used. Batch normalization \citep{Ioffe2015} was applied directly after all the convolutional layers of VGG16. On top of that, dropout regularization \citep{Srivastava2014} was applied in some cases and L2 regularization in others. It turned out that dropout regularization in most cases deteriorated network performance. Conversely, L2 regularization seemed to improve the network performance, albeit with a small margin. In addition, Image Data Augmentation was used as an extra form of regularization. 

To show the effect of Image Data Augmentation, the same convolutional neural network was both trained with and without Image Data Augmentation. In both scenarios, the network was trained for a maximum of 40 epochs, with a learning rate of 0.0001, gradual learning rate decay (0.0001 / 40 = 2.5e-06), momentum of 0.9, and no nesterov. An early stopping criterium was used if the validation loss did not improve in 5 epochs. 

\begin{figure}[h]
	\centering
	\subfloat {%
		\includegraphics[width=0.45\textwidth]{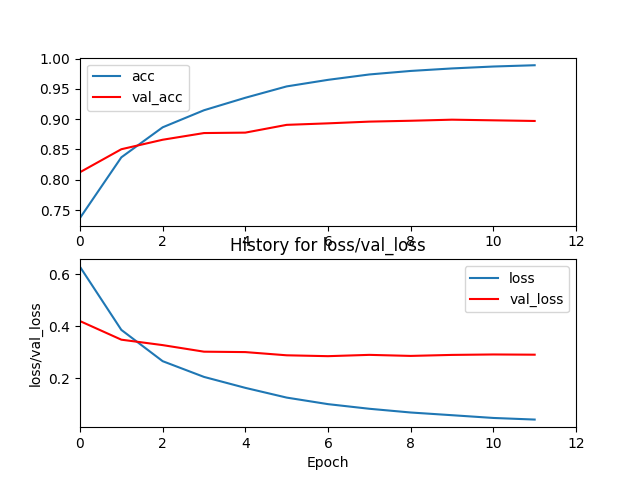}
	}	
	\subfloat {%
		\includegraphics[width=0.45\textwidth]{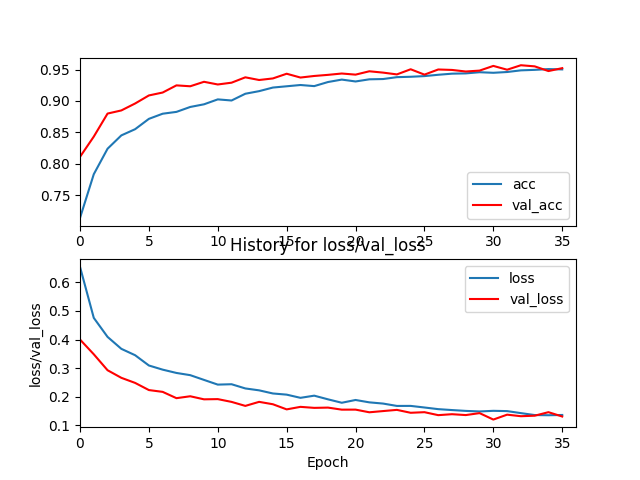}
	}
	
	\caption{Loss and accuracy curves for training and subsequent testing for the same dataset with learning rate = 0.0001, decay = 2.5e-06, momentum = 0.9, nesterov = false. Figure (a) shows the case without image data augmentation, figure (b) shows the case with image data augmentation}	
	\label{fig:with_without_augmentation}
\end{figure}

Figure \ref{fig:with_without_augmentation}a, shows the loss curve for the case without using image data augmentation. It can be seen clearly that within the first epochs the CNN already starts overfitting; the training loss is declining while the validation loss starts increasing after epoch 6. After 12 epochs validation loss/accuracy does not improve and the training is stopped (early stopping). Conversely, in figure \ref{fig:with_without_augmentation}b the validation loss and accuracy curves follow the training loss and accuracy curves much more closely. Training is only stopped after 35 epochs when loss/accuracy stopped improving. 

Summarizing, it was found that when adding image data augmentation, loss and accuracy curves were much smoother than without. In addition, the validation loss and accuracy curves follow the training loss and accuracy curves much closer than without Image Data Augmentation. This means that Image Data Augmentation results in less overfitting.

\subsection{Freezing/Unfreezing layers}
Several experiments were performed with freezing and unfreezing the pre-trained layers up until a certain layer. By freezing the pre-trained layers, they are not changed during fine-tuning. Conversely, when unfreezing layers, the layers are allowed to be trained during fine-tuning. In section \ref{section:fixed_feature_extractor}, it was found that especially the higher level layers of VGG16 may not be suitable for solar panel classification. In the experiments, the networks were therefore at first trained with all convolutional layers frozen. After that, gradually more and more layers were unfrozen until all layers were allowed to be trained. It was found that the networks with the first two convolutional blocks frozen were performing best. The first two convolutional blocks typically contain lower level features like edges, colors and textures. By fixing these features and allowing the higher level features to be retrained, the networks can be adapted to the solar panel classification task without losing the general features that are applicable to all computer vision tasks.

\subsection{Optimizers}
All models were trained at first with Stochastic Gradient Descent (SGD) with different learning rates, momentum, nesterov and learning rate decay configurations. The most successful learning rates were typically small at $1e-4$ and $1e-5$, which is to be expected when using pre-trained models: small learning rates do not change the model weights too drastically from the pre-trained ones. In general it was found that learning rates $1e-4$ and $1e-5$ resulted in the most gradual loss/accuracy curves. In addition, the effect of adding a learning rate scheduler does not seem to have a big effect; the difference between loss/accuracy curves for a fixed learning rated versus gradual learning rate decay with the learning rates mentioned above was small. Likewise, the effect of different momentum values did not seem to have a big effect. Momentum values of 0.7, 0.8 and 0.9 were evaluated. For future cases, the momentum value can be kept at the default value of 0.9. Finally, turning on nesterov seems to lead to more overfitting in all cases. After the experiments with stochastic gradient descent, mostly an RMSprop \citep{Tieleman2012} optimizer was used as that proved more successful and led to slightly higher classification performance. The RMSprop optimizer was used with the same learning rates as stochastic gradient descent ($1e-4$ and $1e-5$) with all other hyperparameters left to the default.

\subsection{Best Networks: Results}	

In the end, it was found that neural network that consisted of all convolutional layers of VGG16 until the last convolutional block (block 5) with one fully connected classification layer was performing best. Both sigmoid and softmax layers were used as classification layer. Performance for both types of classification layers was equivalent. The neural network was trained with all convolutional layers until layer 7 frozen (the first two convolutional blocks), RMSprop with a learning rate of 1e-5, and a L2-regularization on the fully connected layer with strength 0.01 for the sigmoid layer and strength 0.001 for the softmax layer. Adding L2 regularization improved the network performance to some degree. Next to the test set taken from the Heerlen dataset, the model performance was also evaluated on a validation set that was created from a bounding box of the same size as the Heerlen data but moved 100m to the west to prevent it from overlapping with the training/test dataset. The classification metrics for the network with a sigmoid layer can be seen in table \ref{table:classification_performance_vgg16_sigmoid_test} for the test set and table \ref{table:classification_performance_vgg16_sigmoid_validation} for the validation set. It can be seen that the validation set performance is very similar for the negative case, but differs in term of precision for the positive case. With a validation recall of 98\% and a precision of 68\%, the network is able to retrieve most solar panels in the validation set, but also retrieves a lot of false positives as the low precision indicates; this indicates that the network did not learn to identify all solar panels correctly yet. Similarly, the classification metrics for the network with a softmax layer can be seen in table \ref{table:classification_performance_vgg16_softmax_test} for the test set and table \ref{table:classification_performance_vgg16_softmax_validation} for the validation set.

\begin{table}[h!]
    \centering
    \begin{tabular}{c|c|c|c|c}
         & precision & recall & f1-score & support \\
         \hline
           0 & 0.98 & 0.93 & 0.96 & 2473 \\
           1 & 0.85 & 0.96 & 0.90 & 1047 \\
           \hline
   accuracy & & & 0.94 & 3520 \\
   macro avg & 0.92 & 0.95 & 0.93 & 3520 \\
weighted avg & 0.94 & 0.94 & 0.94 & 3520 \\
    \end{tabular}
    \caption{Classification metrics for a fine-tuned VGG16 network with a sigmoid classification layer for the test set}
    \label{table:classification_performance_vgg16_sigmoid_test}
\end{table}

\begin{table}[h!]
    \centering
    \begin{tabular}{c|c|c|c|c}
         & precision & recall & f1-score & support \\
         \hline
           0 & 1.00 & 0.92 & 0.96 & 11384 \\
           1 & 0.68 & 0.98 & 0.81 & 1968 \\
        \hline
   accuracy & & & 0.93 & 13352 \\
   macro avg & 0.84 & 0.95 & 0.88 & 13352 \\
weighted avg & 0.95 & 0.93 & 0.93 & 13352 \\
\end{tabular}
    \caption{Classification metrics for a fine-tuned VGG16 network with a sigmoid classification layer for the validation set}
    \label{table:classification_performance_vgg16_sigmoid_validation}
\end{table}

\begin{table}[h!]
    \centering
    \begin{tabular}{c|c|c|c|c}
         & precision & recall & f1-score & support \\
         \hline
           0 &  0.96 & 0.97 & 0.97 & 2471 \\
           1 &  0.94 & 0.91 & 0.92 & 1049 \\
        \hline
   accuracy & & & 0.96 & 3520 \\
   macro avg & 0.95 & 0.94 & 0.95 & 3520 \\
weighted avg & 0.96 & 0.96 & 0.96 & 3520 \\
 samples avg & 0.96 & 0.96 & 0.96 & 3520 \\
    \end{tabular}
    \caption{Classification metrics for a fine-tuned VGG16 network with a softmax classification layer for the test set}
    \label{table:classification_performance_vgg16_softmax_test}
\end{table}

\begin{table}[h!]
    \centering
    \begin{tabular}{c|c|c|c|c}
         & precision & recall & f1-score & support \\
         \hline
           0 &  0.99 & 0.97 & 0.98 & 11384 \\
           1 &  0.85 & 0.96 & 0.90 & 1968 \\
         \hline
   accuracy & & & 0.97 & 13352 \\
   macro avg & 0.92 & 0.96 & 0.94 & 13352 \\
weighted avg & 0.97 & 0.97 & 0.97 & 13352 \\
\end{tabular}
    \caption{Classification metrics for a fine-tuned VGG16 network with a softmax classification layer for the validation set}
    \label{table:classification_performance_vgg16_softmax_validation}
\end{table}

All in all, the results on the test sets are quite equivalent. It can be seen that both fine-tuned networks outperform the feature extractors in section \ref{section:fixed_feature_extractor} by a great margin. The classification performance for the negative case has been improved by 4\%-10\% while the classification for the positive classes has been increased by at least 10\% but in some cases even as much as 30\%. This again confirms our previous conclusion that the higher-level features learned by VGG16 on the ImageNet \citep{ILSVRC15} are not that suitable for use with aerial images. They do perform a good starting point for further fine-tuning as the classification performance in table \ref{table:classification_performance_vgg16_sigmoid_test} and \ref{table:classification_performance_vgg16_softmax_test} show. While the classification performance for the sigmoid and softmax classification layer are somewhat equivalent, the softmax classification layer outperforms the sigmoid classification layer on the validation set. The network with the softmax classification layer thus generalizes better. An added benefit of the neural network with a softmax classification layer, is that the softmax layer makes it possible to create class activation maps \citep{zhou2015cnnlocalization, DBLP:journals/corr/SelvarajuDVCPB16}. Class activation maps provide visualizations of the regions in an image that were activating strongly and influenced the final classification output greatly. Class activation maps, as such, are a great tool to look inside the network and debug false classifications.

\section{Phase IV: Blind Test on NRW and Limburg}
\label{section:phase4_blind_test}

In this test we took two areas, one from NRW in Germany and one from Limburg in the Netherlands, and then selected 10 000 locations out of which we performed a classification by either mean of a Inception 10649 imResNet2 or using object detection. The validation sample is taken with its natural distribution without any rebalancing in order to observe how the algorithm performs in real settings. In the case of the NRW data, this shows the ability of the model to generalize on images coming from the same population of aerial images. In the case of Limburg, the validation shows the ability of the model to be transferred to a different population of aerial images. We therefore report their performance and discuss qualitatively the types of mistakes in the two regions. We also discuss the ability of the deep learning model to detect solar panels that are not contained in the registries (LANUV case), as this gives us the final indication on whether or not there is an added value in using computer vision for the task of completing the registries.

\begin{figure}[htp!]
 \centering
 \subfloat[Bonn]{\includegraphics[width=0.45\textwidth]{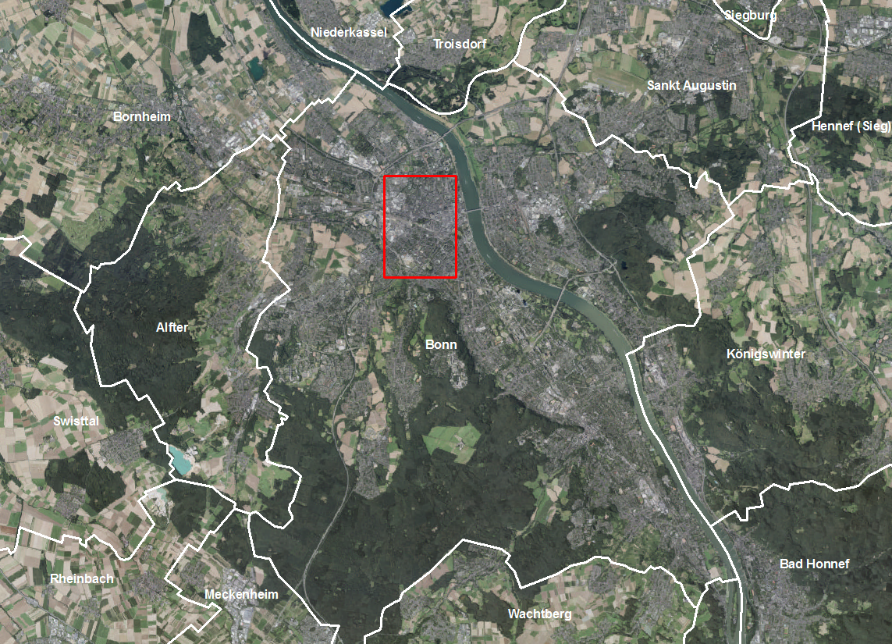}\label{fig:bonnarea}}
 \subfloat[Düren]{\includegraphics[width=0.45\textwidth]{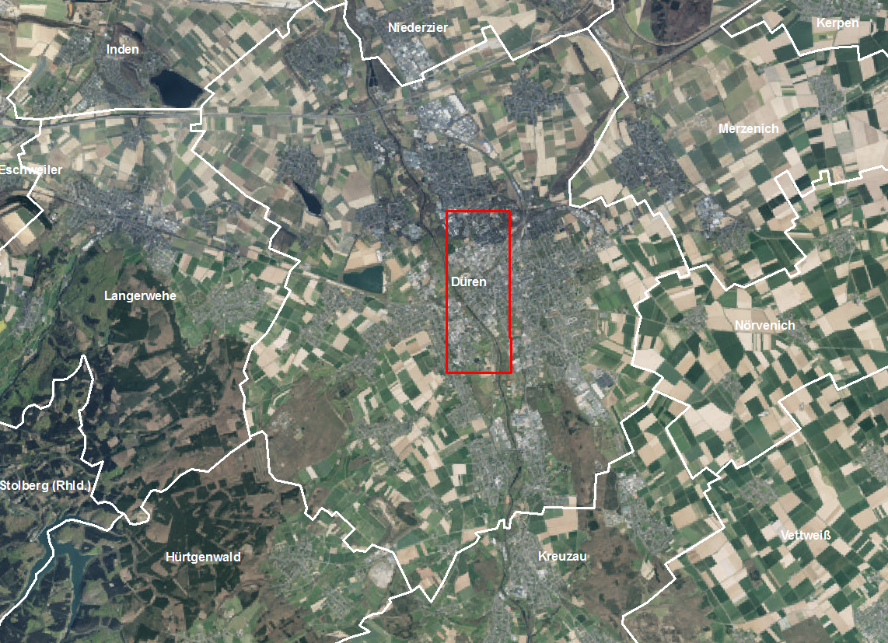}\label{fig:durenarea}}
 \caption{Bonn and Düren areas selected for the testing. }
 \label{fig:area}
\end{figure}

NRW was provided with a csv with 10 000 classifications of aerial images. Those classifications are the result of a learning algorithm which was created to find out whether there is a solar panel in the image or not. In the validation process those predicted classifications had to be checked. Figure \ref{fig:area} shows the two areas that we considered for the analysis.

\begin{figure}[htp!]
 \centering
 \subfloat[Bonn]{\includegraphics[width=0.45\textwidth]{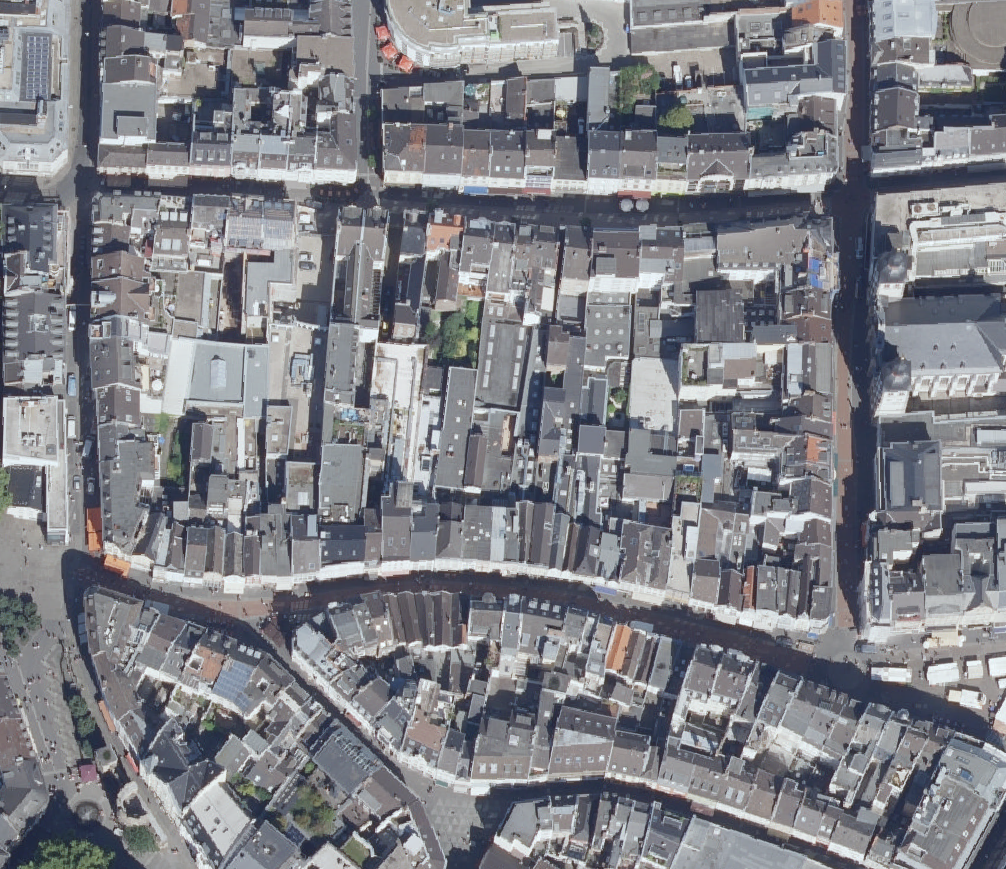}\label{fig:bonsett}}
 \subfloat[Düren]{\includegraphics[width=0.45\textwidth]{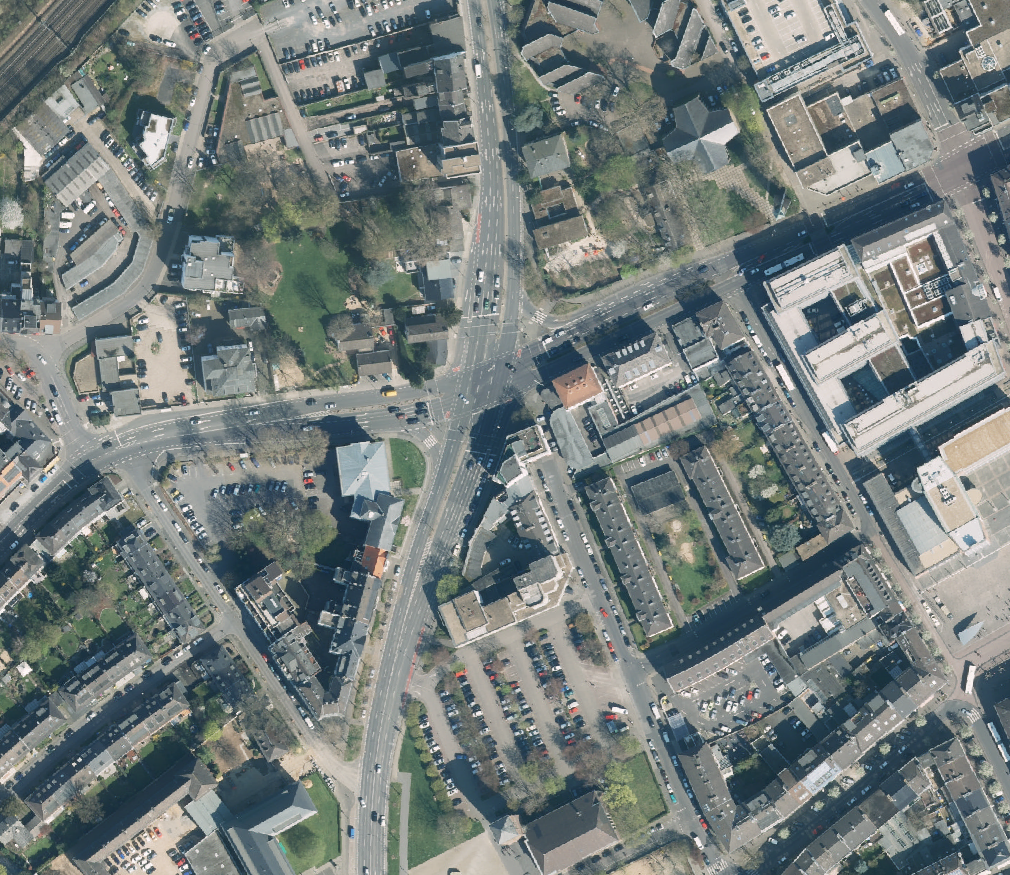}\label{fig:dursett}}
\caption{Bonn and Düren areas selected for the testing. }
\label{fig:sett}
\end{figure}

One important aspect to consider is that the two areas differ a lot in terms of settlement structure, for an example see figure \ref{fig:sett}.
Bonn results in a much denser urban area than Düren and this has an effect on the ability of the deep learning models used to distinguish false positives.

The expert evaluation was carried out with ArcGIS. The software was used to build 50m $\times$ 50m bounding boxes around each given coordinate pair. The predictions from the algorithm have then been visually checked in ArcGIS against the dop10 WMS-service. Every tile has been classified (as positive or negative) for solar-panels. Afterwards the predictions from the model have been compared with the results from the visual classification.  

\begin{figure}[htp!]
 \centering
 \subfloat[Bonn, Prediction Grid]{\includegraphics[width=0.45\textwidth]{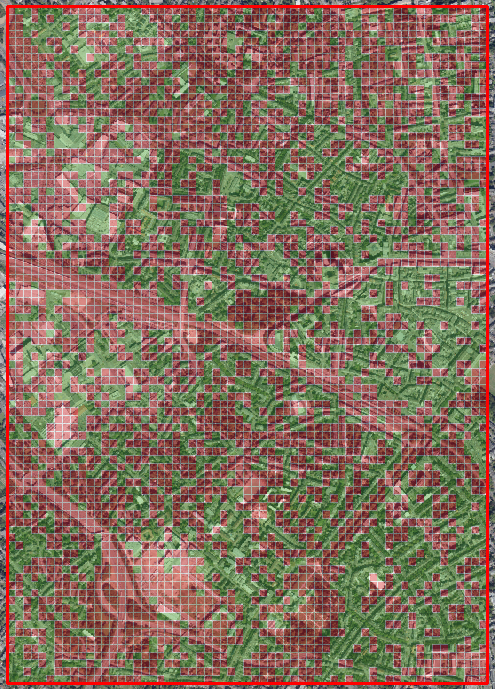}\label{fig:bonngrid}}
 \subfloat[Düren]{\includegraphics[width=0.45\textwidth]{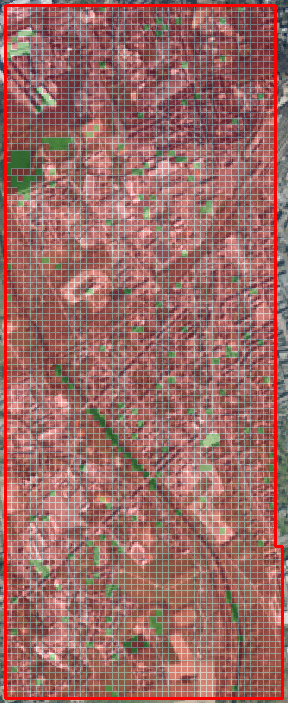}\label{fig:durgrid}}
 \label{img:grid}
  \caption{Bonn and Düren Classification Results, Visual Summary. }
 \end{figure}

\begin{table}[htp!]
\begin{tabular}{|l|l|l|}
\hline
Result                                                     & Amount        & \begin{tabular}[c]{@{}l@{}}\%\\   of total\end{tabular} \\ \hline
\begin{tabular}[c]{@{}l@{}}False\\   Negative\end{tabular} & 67            & 0.67                                                    \\ \hline
\begin{tabular}[c]{@{}l@{}}False\\   Positive\end{tabular} & 2277          & 22.77                                                   \\ \hline
\begin{tabular}[c]{@{}l@{}}True\\   Negative\end{tabular}  & 7514 (7515)   & 75.14                                                   \\ \hline
\begin{tabular}[c]{@{}l@{}}True\\   Positive\end{tabular}  & 142           & 1.42                                                    \\ \hline
Total                                                      & 10000 (10001) & 100                                                     \\ \hline
\end{tabular}

\caption{Results: Bonn and Düren combined.}
\label{tab:bonndu}

\end{table}

\begin{table}[htp!]
\begin{tabular}{|l|l|l|}
\hline
Result                                                     & Bonn        & \begin{tabular}[c]{@{}l@{}}\%\\   of total\end{tabular} \\ \hline
\begin{tabular}[c]{@{}l@{}}False\\   Negative\end{tabular} & 14          & 0.27                                                 \\ \hline
\begin{tabular}[c]{@{}l@{}}False\\   Positive\end{tabular} & 2100        & 40.03                                              \\ \hline
\begin{tabular}[c]{@{}l@{}}True\\   Negative\end{tabular}  & 3027 (3028) & 57.70                                              \\ \hline
\begin{tabular}[c]{@{}l@{}}True\\   Positive\end{tabular}  & 105         & 2.00                                              \\ \hline
Total                                                      & 5246 (5247) & 100  \\
\hline

\end{tabular}
\caption{Results: Bonn specific results.}
\label{tab:bonn}

\end{table}

\begin{table}[htp!]
\begin{tabular}{|l|l|l|}
\hline
Result                                                     & Düren & \begin{tabular}[c]{@{}l@{}}\%\\   of total\end{tabular} \\\hline
\begin{tabular}[c]{@{}l@{}}False\\   Negative\end{tabular} & 53    & 1.11\\ \hline
\begin{tabular}[c]{@{}l@{}}False\\   Positive\end{tabular} & 177   & 3.72\\ \hline
\begin{tabular}[c]{@{}l@{}}True\\   Negative\end{tabular}  & 4487  & 94.38\\ \hline
\begin{tabular}[c]{@{}l@{}}True\\   Positive\end{tabular}  & 37    & 0.78\\ \hline
Total                                                      & 4754  & 100  \\         
\hline
\end{tabular}
\caption{Results: Düren specific results.}
\label{tab:duren}
\end{table}

One remarkable aspect of this validation is that in Bonn we have a higher false positive rate than in Düren. We suspect that this is caused by the higher urban density in Bonn that tends to confuse the deep learning model used with features that resemble solar panels. 

The hypothesis is that most of the false positives are caused by structures that are similar to solar panels or that present windows or objects that are subdivided in a grid structure, that would make them difficult to distinguish from real solar panels. If we perform a quick analysis of the errors we can see that this is the case for most of the errors.

\begin{figure}[htp!]
 \centering
 \subfloat[]{\includegraphics[width=0.25\textwidth]{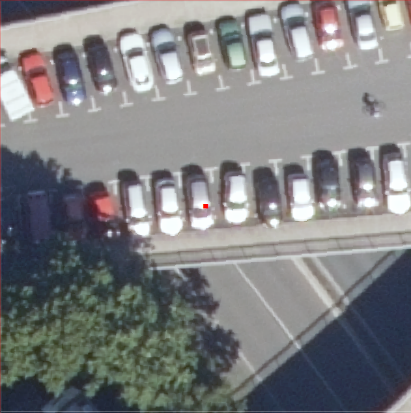}\label{fig:fp1}}
 \subfloat[]{\includegraphics[width=0.25\textwidth]{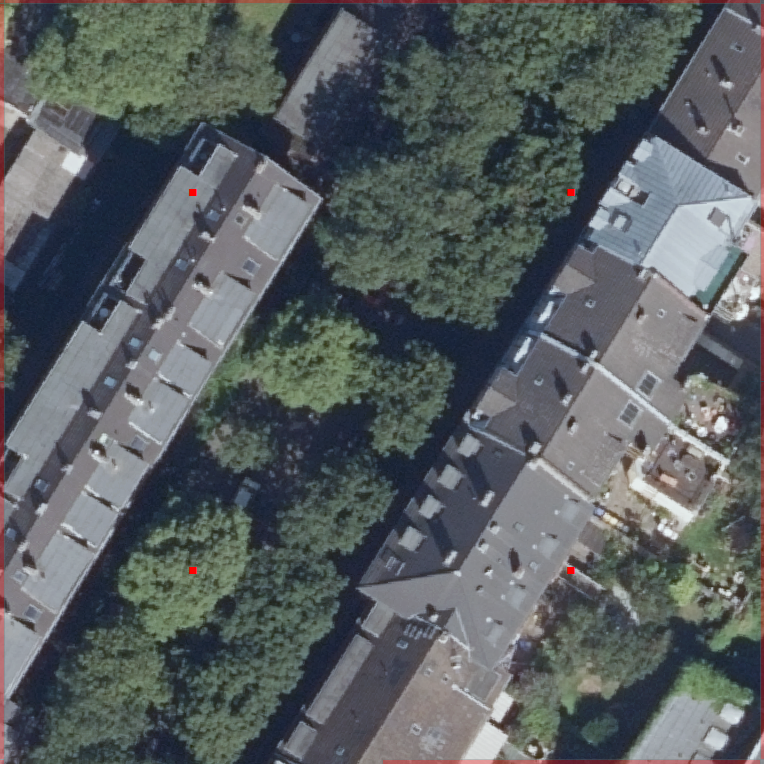}\label{fig:fp2}}
 \subfloat[]{\includegraphics[width=0.25\textwidth]{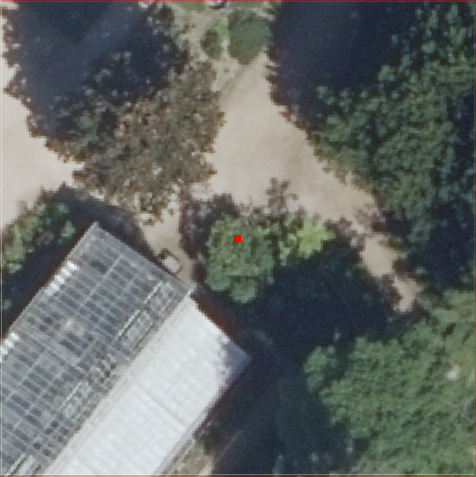}\label{fig:fp3}}
 \subfloat[]{\includegraphics[width=0.25\textwidth]{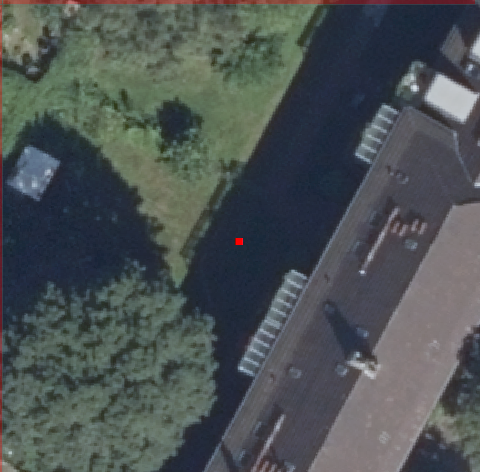}\label{fig:fp4}}

 \label{img:sett}
  \caption{False positives examples. }
\end{figure}

\subsection{Newly Discovered Solar Panels in NRW}

\label{section:newly_discovered_solar_panels}

One important part of the analysis is whether or not the deep learning models can detect new solar panels as the main goal is to achieve a better quality registry. 

\begin{figure}[htp!]
 \centering
 \subfloat[Bonn]{\includegraphics[width=0.45\textwidth]{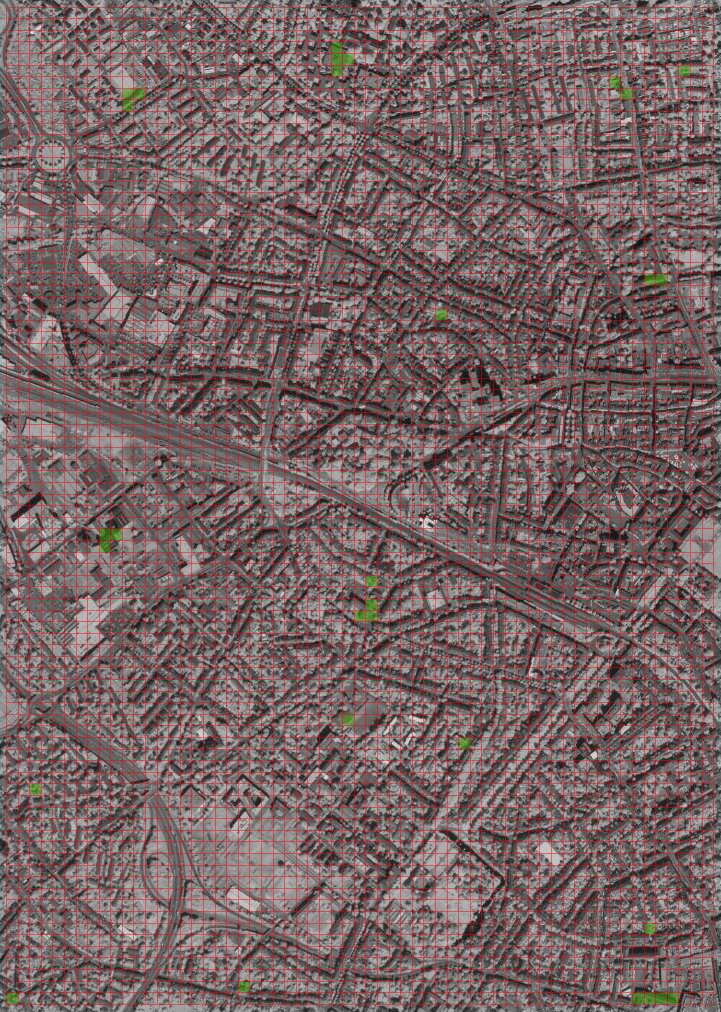}\label{fig:bonnareaa}}
 \subfloat[Düren]{\includegraphics[width=0.45\textwidth]{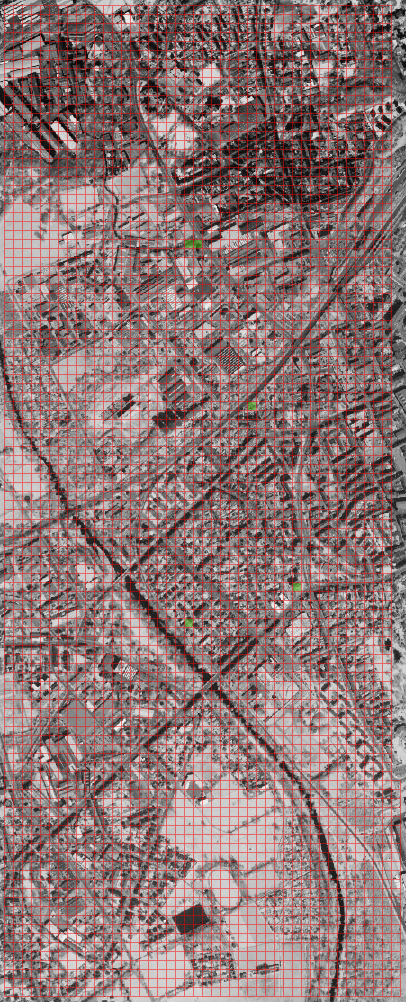}\label{fig:durenareab}}
  \caption{Bonn and Düren areas selected for the testing. }
  \label{fig:orthosab}
\end{figure}

Some adaptation from the LANUV registry was required in order to identify solar panels that are not in the registry: the coordinates from the LANUV are point coordinates which represent only the address where a solar panel is registered. To compare the detected solar panels (true positives) with the registered solar panels from the LANUV, it was necessary to change the address point-features to polygon-features, because otherwise not all positive tiles would have been taken into account. The creation of the polygon-features was done manually by sighting each address together with the corresponding orthophotos (see Fig. \ref{fig:orthosab}).

\begin{table}[htp!]
\begin{tabular}{|l|l|l|}
\hline
Result                                                   & Amount & \begin{tabular}[c]{@{}l@{}}\%\\   of total\end{tabular} \\ \hline
LANUV                                                    & 107    & 75,35                                             \\ \hline
\begin{tabular}[c]{@{}l@{}}not\\   in LANUV\end{tabular} & 35     & 24,65                                             \\ \hline
Total                                                    & 142    & 100                                                     \\ \hline
\end{tabular}

\caption{Newly discovered Solar Panels in Bonn and Düren, Summary.}
\label{tab:bondunew}

\end{table}

Table \ref{tab:bondunew} reports the newly found solar panels for both Bonn and Düren
combined. With the classification-algorithm about 25\% of all true Solar panels could be found, which have not been registered in the LANUV data. In Bonn the amount is larger, but one has to reconsider the high false positive rate for the algorithm in that area. Also in Düren there were 5 Solar panels, which were not in the LANUV data.

As an additional test, we tried object detection on the images provided by NRW. For the purpose of training the Mask R-CNN object detection algorithm, we used the 5000 images data set comprising the polygons enclosing the solar panels in the images, described in section \ref{section:datasets}.

The classification report of both object detection with Mask RCNN and InceptionV2 are reported below.

\begin{table}[h!]
    \centering
    \begin{tabular}{c|c|c|c|c}
         & precision & recall & f1-score & support \\
         \hline
           0 &  0.99 & 0.87 & 0.93 & 9792 \\
           1 &  0.10 & 0.66 & 0.17 & 209 \\
        \hline
  accuracy &  &  & 0.86 & 1065 \\
   macro avg & 0.54 & 0.76 & 0.55 & 10001 \\
weighted avg & 0.97 & 0.86 & 0.91 & 10001 \\
    \end{tabular}
    \caption{Classification metrics for a Mask RCNN object detection model on a NRW data set of Düren and Bonn.}
    \label{tab:inc_mrcnn}
\end{table}

\begin{table}[h!]
    \centering
    \begin{tabular}{c|c|c|c|c}
         & precision & recall & f1-score & support \\
         \hline
           0 &  0.99 & 0.83 & 0.90 & 9792 \\
           1 &  0.07 & 0.62 & 0.13 & 209 \\
        \hline
  accuracy &  &  & 0.82 & 1065 \\
   macro avg & 0.54 & 0.73 & 0.52 & 10001 \\
weighted avg & 0.97 & 0.82 & 0.89 & 10001 \\
    \end{tabular}
    \caption{Classification metrics for a InceptionV2 model on a NRW data set of Düren and Bonn.}
    \label{tab:inc_incept}
\end{table}

Tables \ref{tab:inc_mrcnn} and \ref{tab:inc_incept} show a comparison concerning prediction, recall and f1-score with respect to classification.

The object detection algorithm is predictive with respect to the blind test, but its performance remains comparable to the one of InceptionV2. This may be caused by a number of factors, such as for example the fact that we train with less images than with the classification, or the complexity of the scenery in the aerial image. Given the additional complexity of annotating images with polygons and the longer training times of object detection algorithms, a final decision of using classification algorithms has been made in our study.    

\subsection{Phase IVb: NRW model applied to Limburg Data}
\label{section:phase_4b_nrw_to_limburg}

As an additional test, we decided to test the transferability of the model to DOP10 data of the Limburg area, focusing on the images collected around the city of Heerlen. The tables below provide a summary of the results of three approaches: Mask R-CNN (see table \ref{fig:summary_results_nrw_heerlen_mask_rcnn}), InceptionResNetV2 (see table \ref{fig:summary_results_nrw_heerlen_inception}), and a combination of the two (see table \ref{fig:summary_results_nrw_heerlen_maskrnn_inception}). The green cell values in the tables represent the best results for a certain class (positive/negative) and metric type (precision, recall, or f1-score).

\begin{table}[h!]
    \centering
    \begin{tabular}{c|c|c|c|c}
         & precision & recall & f1-score & support \\
         \hline
           0 &  0.93 & \textcolor{green}{0.59} & \textcolor{green}{0.72} & 15991 \\
           1 &  0.39 & 0.86  & 0.54 & 4913 \\
         \hline
   accuracy & & & 0.66 & 20904 \\
   macro avg & 0.66 & 0.73 & 0.63 & 20904 \\
weighted avg & 0.81 & 0.66 & 0.68 & 20904 \\
        \hline  
\end{tabular}
    \caption{Transfer Learning Results for Mask R-CNN Object Detection model on the Heerlen area.}
    \label{fig:summary_results_nrw_heerlen_mask_rcnn}
\end{table}

\begin{table}[h!]
    \centering
    \begin{tabular}{c|c|c|c|c}
         & precision & recall & f1-score & support \\
         \hline
           0 &  0.95 & 0.58 & \textcolor{green}{0.72} & 15991 \\
           1 &  \textcolor{green}{0.40} & 0.90  & \textcolor{green}{0.55} & 4913 \\
         \hline
   accuracy & & & 0.66 & 20904 \\
   macro avg & 0.67 & 0.74 & 0.64 & 20904 \\
weighted avg & 0.82 & 0.66 & 0.68 & 20904 \\
        \hline  
\end{tabular}
    \caption{Transfer Learning Results for the InceptionResNetV2 classification model on the Heerlen area.}
    \label{fig:summary_results_nrw_heerlen_inception}
\end{table}

\begin{table}[h!]
    \centering
    \begin{tabular}{c|c|c|c|c}
         & precision & recall & f1-score & support \\
         \hline
           0 &  \textcolor{green}{0.98} & 0.42 & 0.58 & 15991 \\
           1 &  0.34 &\textcolor{green}{0.97}  & 0.50 & 4913 \\
         \hline
   accuracy & & & 0.55 & 20904 \\
   macro avg & 0.66 & 0.69 & 0.54 & 20904 \\
weighted avg & 0.83 & 0.55 & 0.56 & 20904 \\
        \hline  
\end{tabular}
    \caption{Transfer Learning Results for the Mask R-CNN or InceptionResNetV2 combined model on the Heerlen area.}
    \label{fig:summary_results_nrw_heerlen_maskrnn_inception}
\end{table}

It is possible to observe from the results in the tables above that the classifier and object detectors developed on the NRW data seem to retain the same statistics as in NRW concerning the F1-score, meaning that the regions of NRW and Limburg seem to share similar architectures. Nonetheless, despite presenting a similar F1-score we can see that the precision is lower and the recall higher than in the NRW region. We believe that this difference is due to the fact that the density of panels is a lot higher in Heerlen data (almost 20\% of the images contained a solar panel) than in NRW data (only 2\% of the testing images contained solar panels).

\subsection{Phase IVc: Limburg model applied to NRW data}

As a last test, we also evaluated the model that was trained on the Heerlen region on NRW data. The same random sample as in section \ref{section:phase4_blind_test} was taken and re-annotated, because the input size for the Heerlen network was smaller than that of the NRW network. After annotating, only a small amount of solar panels was left in the original data. In the end, the validation set consisted of 9935 images without solar panels and 107 images with solar panels (roughly half the amount of solar panels as with $330\times330$ pixel images). The best performing Heerlen network, the one with the softmax layer, was applied on this dataset without any retraining. Table \ref{table:classification_performance_vgg16_softmax_test_heerlen_nrw} shows the performance of the Heerlen network on NRW data.

\begin{table}[h!]
    \centering
    \begin{tabular}{c|c|c|c|c}
         & precision & recall & f1-score & support \\
         \hline
           0 & 1.00 & 0.96 & 0.98 & 9828 \\
           1 & 0.15 & 0.69 & 0.25 &  107 \\
        \hline
   accuracy & & & 0.96 & 9935 \\
   macro avg & 0.58 & 0.83 & 0.61 & 9935 \\
weighted avg & 0.99 & 0.96 & 0.97 & 9935 \\ 
 \end{tabular}
    \caption{Classification metrics for a fine-tuned VGG16 network with a softmax classification layer, trained on Heerlen and applied to NRW}
    \label{table:classification_performance_vgg16_softmax_test_heerlen_nrw}
\end{table}

When comparing the results in table \ref{table:classification_performance_vgg16_softmax_test_heerlen_nrw} with the validation set for the Limburg region in table \ref{table:classification_performance_vgg16_softmax_validation}, we can see that the precision, recall, and f1-score for the negative cases stay roughly the same. The precision, the recall, and f1-score for the positive case all drop significantly. All in all, with a recall of 69\% the network still manages to identify a reasonable amount of solar panels. The precision of only 15\% however indicates that there is a substantial amount of false positives to deal with. The lower recall and precision may result from different building styles and also different types of solar panels. It was found that in the sampled dataset of NRW, most houses have pitched roofs, whereas in Netherlands a lot more flat roofs were present. Moreover, some types of solar panels have been seen that we have not yet encountered in the Dutch aerial pictures.

\chapter{Conclusions}
\label{chapter:outlook}

At the outset of the project, we aimed to address the following questions:
\begin{enumerate}
    \item What is the usability of different types of images (i.e. aerial and satellite) and minimally required image resolution for detecting solar panels?
    \item Which method is best suited for detecting solar panels?
    \item Is it possible to develop a harmonized method across EU member states?
\end{enumerate}
We will address each of these questions in turn in the following subsections.

\section{What is the usability of different types of images (i.e. aerial and satellite) and minimally required image resolution for detecting solar panels?}

In the beginning of the project, we identified two types of possible data sources to be used for the detection of solar panels: aerial images and satellite images. For the aerial images, we evaluated three different resolutions:
\begin{itemize}
    \item DOP10 or $10x10cm$,
    \item DOP20 or $20x20cm$,
    \item DOP25 or $25x25cm$.
\end{itemize}

We started out with DOP20 images for the North Rhine Westphalia region and DOP25 images for the Limburg region. We decided that to make the granularity of the classifier as fine-grained as possible, with still having enough context, to use images of $75 \times 75$ pixels as input. $75 \times 75$ pixels amounts to $15 \times 15m$ in DOP20 case and $18\frac{3}{4} \times 18\frac{3}{4}m$ in the DOP25 case. With such a resolution, every image contained, \emph{at most}, a couple of houses. While a small resolution such as $75\times75$ provides us with a fine granularity, most convolutional neural networks trained on ImageNet work on images that have a resolution larger than $200 \times 200$ pixels; to work with the smaller resolution the state-of-the-art networks therefore need to be adapted. An alternative would be to increase the resolution of the images, and therefore the area covered, to work with standard out-of-the-box algorithms. When the area is increased, the granularity of the classification algorithm is decreased as there are now more houses on the image (because there is only one classification given per image, this classification now holds for more houses than in the more granular case). 

A first evaluation with low resolution data trained the state-of-the-art convolutional neural networks without making any changes to their structure. While the performance on the test set was high, performance on the validation set showed issues with generalizability. It is expected that the low performance on the validation set was caused by the disparity of the resolution expected by the full models ($226\times226$ pixels) as opposed to the resolution that they were trained on  ($75\times75$ pixels). To improve performance, we therefore evaluated the DOP10 dataset, where a similar resolution in meters resulted in images with pixel dimensions that are much more similar to the input dimensions of most out-of-the-box networks. 

From the results it becomes clear that, the validation performance and generalization of the convolutional neural networks is higher than that on the the DOP25 dataset. Not only is the input resolution closer to that of the standard convolutional neural networks, the DOP10 datasets also offer more detail and therefore make the distinction between pictures with solar panels and without solar panels easier. This however does not mean that DOP20 and DOP25 resolutions cannot be used for solar panel classification. We expect that for a large quantity of solar panels the resolution will be sufficient, but that the lower detail level will make it more difficult to give a proper classification in some cases. In this respect, it was found that in the low resolution pictures it is more difficult to distinguish certain features, like for example roof top windows, from solar panels. These features differences are much more apparent in the high resolution images.

Moreover, we expect that with convolutional neural networks specifically adapted to the lower input resolution, performance will be better. In addition, the neural network weights learned on the high resolution images could be used for transfer learning on the lower resolution images. In this case, it is especially interesting to see if certain cases fail to work when using low resolution images or if they are sufficient to deal with all detection cases after all. It needs to be seen which resolution is still sufficient to still detect solar panels properly. In \cite{malof2015} for example, DOP30 aerial images are successfully used to identify solar panels. In contrast to that, we found that satellite images with a resolution of $50\times50cm$ do not provide enough detail to discern solar panels. Hence, more research is needed on (1) the applicability of the lower resolution pictures, especially which solar panels can be detected and which cannot, (2) what happens to network performance when image resolution decreases, and (3) the convolutional neural networks that should be used with low resolution pictures and how existing networks should be adapted.

\section{Which method is best suited for detecting solar panels?}

As we have seen in the chapter \ref{chapter:results}, convolutional neural networks were used in three ways for solar panel detection. First, we used convolutional networks as feature extractor. Second, we fine-tuned convolutional neural networks to use them as an image classifier. Last, we used convolutional neural networks as an object detector.

Out-of-the-box performance of neural networks trained on ImageNet, as was the case when neural networks were used as feature extractor, left much to be desired. While especially successful in identifying the negative classes, the positive performance topped at 73 \% precision, 67\% recall and 70\% F1-score. The lower performance on the positive classes seems to indicate that higher level features in the neural network are not that applicable to solar panel classification. This becomes even more apparent when fine-tuning the convolutional neural networks on the image classification task. By retraining the higher level layers, but keeping the lower level ones the same, we were able to obtain much higher classification performance for the positive as well as the negative cases.

For the Heerlen area, two neural networks were found to have the best performance. Both networks were based off pre-trained VGG16 convolutional layers and added the classification layers directly on top. Next to that, the first two convolutional blocks were frozen, whereas the rest of the convolutional layers were retrained. We trained one network with a sigmoid layer and one with a softmax layer. While equivalent in test performance, the network with the softmax layer outperformed the one with the sigmoid layers in terms of precision for the positive class. 

Especially, the positive precision dropped significantly for the network with sigmoid layer. The lower precision (68\%) in combination with a good recall (98\%) indicates that the sigmoid network is able to retrieve a good amount of the solar panels but cannot distinguish between positive and negative classes that well. Because of that, a significant amount of false positives (32\%) are returned by the network. While the network learned the correct features to identify a large part of the solar panels in the validation set, the differences between which features lead to a positive and which to a negative classification should be more pronounced.

In contrast, the network with the softmax layer had a good recall, precision, and f1-score for all classes. With a precision of 85\%, the network is able to identify much more of the solar panels correctly than the network with the sigmoid layer. However, with a recall of 96\% it is able to retrieve less solar panels from the total population than the sigmoid network. However, the difference in positive recall is small (2\%) and overall performance for the softmax network is much better. In both cases, we think that adding more examples out of the population of false positives will lead to a better precision. In addition, we think that both networks form a good starting point for further training with additional images. The high recall value for both networks also makes them suitable to be used in pre-screening step for human observers: by using the networks to pre-classify the total population of images, a smaller subset can be filtered out that can be manually checked. The networks can also be used as an additional quality assurance for register data by comparing the predictions of the network with the values given in the register.

For what concerns NRW data, we approached the problem by trying both image classification and object detection, noticing that image classification is better at identifying images containing solar panels. For a later development, object detection may still be useful to estimate the area covered by the solar panels, nonetheless we decided to adopt the InceptionV2 model for the experimentation on the basis that in testing and validation it had the best precision and recall for both negative and positive cases. 

What is possible to notice from training, testing and validation, is that there is an identifiable pattern in the images that allows to produce trained models that would certainly go better than a random baseline. But it is also clear that the problem remain challenging as the false positive rate is difficult to reduce. In NRW we decided to use two different areas for the validation of the results, Düren and Bonn. We discovered that the different architectural styles of the cities have an impact on how well the models can detect solar panels, with Düren presenting a much better result than Bonn for what concerns the false positive rate ( 40\% vs 3.7\%), despite having a similar density of solar panels. Despite presenting this difference in terms of false positives, in both the areas of Bonn and Düren, the InceptionV2 model identified solar panels that were not previously present in the LANUV data set. This effectively means that the developed models can already be used to improve the quality of the current registries and saving time in terms of manual annotation of images.

The results of cross validation on distinct geographical areas within the same aerial picture are mixed. The network trained on the Heerlen region, and validated on an area close to the training area, especially showed a drop in the precision for the positive case while other performance metrics stayed roughly similar. The differences for the North Rhine Westphalia (NRW) area were more pronounced. A possible cause for the larger gap in test and validation performance, is the bigger distances between training area and validation area in NRW. In addition, the NRW network was trained on a much larger geographical region. The network was therefore confronted with a much larger variety in geographical areas, urban planning, and architecture. Moreover, these differences in geographical areas, urban planning, and architecture may vary more from country to country. While the performance of the Heerlen network was quite stable, it still needs to be seen how well the network performance scales to areas that are further apart and differ more in terms of building styles and urbanization. A next step should evaluate network performance on areas in different provinces of the Netherlands.

To conclude, we see that fine-tuning convolutional neural networks to the domain of aerial images and in particular solar panel detection, performs much better than using the network as-is and as a feature extractor. The differences between the performance of networks used for image classification on the one hand, and object detection on the other, are less pronounced. We do see that in both cases network performance benefits from using transfer learning, and thus pre-trained weights on ImageNet. For future work, we therefore advise to use fine-tuning and transfer learning to train new networks on aerial images. The networks we trained for the DeepSolaris project, could be a good starting point for transfer learning in this domain. 

\section{Is it possible to develop a harmonized method across EU member states?}

As it currently stands, the issue of detecting solar panels with deep learning models happens to be challenging in the European landscape, in particular because of the change in architecture and style between different countries and even between different cities in the same country. We cross-validated networks trained on the NRW area on the Heerlen area, and vice versa. In both cases, overall network performance dropped with validation in cross-border settings. These results are consistent with the research presented in \citep{Wang2017}, which presents the results of a cross-site evaluation between two cities in California. Without retraining the model trained on one city with data from the other city, network performance also dropped significantly as was the case in our cross-border evaluation. It needs to be seen how network performance develops when examples from another geographical region are added to the training set. Another interesting aspect to investigate is that the drop was not symmetrical, the network trained on NRW data managed to retain a precision of 40\% on the Limburg/Heerlen data and an F1-score of 50\% indicating that the network can generalize across NRW and Limburg, although it has to be noted that the density of solar panels in the Limburg region was way higher than in the NRW region. Another reason for this difference may be that the variability of buildings in NRW is higher than in Limburg, which would allow the NRW algorithm to generalize better than the one in Limburg. 

While performance dropped, both networks still have predictive power and a reasonable recall. To improve performance, the networks may need to be retrained to deal with specific variations in building style and solar panel types. Again, by adding samples from false positives and false negatives, network performance in general may be improved. This is something that should be explored in further work.

While network performance drops significantly in cross-border settings, it seems that the networks still have predictive power in these scenarios. In any case, the method described for annotating images, training and fine-tuning existing neural networks, and evaluating them on other geographical areas of the same country, is applicable to countries across Europe. A possible way to improve the network performance, is to retrain them on specific geographical areas. To create networks that work in a cross-border setting it should be evaluated whether networks trained with images sampled from several countries, fare better than the networks we described here.

\section{Conclusion}

In this project, we trained and evaluated a variety of deep learning algorithms on a range of datasets. We created and labeled datasets for two different geographical regions spread across two different countries. To this cause, software was developed to pre-process aerial images and cut them down to sizes that are manageable by the deep learning algorithms. Moreover, annotation software was developed to help with the task of manually labeling large amounts of images. The cross-border dataset that we developed and labeled consists of 80000 images labeled for image classification and 5000 images labeled for object detection. Specifically for the classification task in NRW a sample was collected out of 50000 images, for which 10000 images with solar panels and 10000 images without solar panels were used for the pure classification task, in addition 5000 positive images were annotated with the polygons containing the solar panels to apply object detection with Mask RCNN. 

It was demonstrated that convolutional neural networks indeed can be used to detect solar panels in aerial images, that these networks are predictive for the task, but that validation results across geographical areas are less outspoken. Especially, the false positive rate stands in the way of creating a fully automatic solution. To improve network performance, we can either add more data or add data in a smarter way, but it still needs to be seen if the false positive rate can be improved sufficiently to provide a full automatic solution. In specific, the small amount of images with solar panels in comparison to the amount of images without solar panels, will demand a model with very high classification accuracy.

Convolutional neural networks are however suitable to automate a part of the task of solar panel detection. With the networks presented in this report, we are able to pre-process large datasets and identify large quantities of solar panel candidates that can then be further processed by human observers. Also, negative precision and recall were sufficient to already discard a large amount of pictures as containing no solar panels. In this way, the networks can greatly reduce manual labor as only parts of the aerial image have to be considered. In addition, the current networks can already be used to improve register quality, as we have seen in section \ref{section:newly_discovered_solar_panels}, the accuracy of the registry could be improved by adding a 25\% to 30\% of previously unknown solar panels. However, to use convolutional neural networks to produce official statistics, there are still some steps to be taken. We will discuss possible directions for further work in chapter \ref{chapter:further_work}.

\chapter{Further Work}
\label{chapter:further_work}
In this section, we will present several suggestions for further work. These suggestions vary from suggestions about the network itself in sections \ref{section:causes_classifier_performance}, \ref{section:improving_classifier_performance}, and \ref{section:one_network}, minimizing the effects of class imbalance in section \ref{section:minimize_effect_class_imbalance}, to evaluate network performance, developing further models (section \ref{section:model_zoo}) and datasets (section \ref{section:cross_border_dataset}), to comparing the model predictions to the register data (section \ref{section:comparison_with_register_data}). 

\section{Understand the causes behind the deterioration of the network performance}
\label{section:causes_classifier_performance}

We have seen that network performance drops when training neural networks in a certain geographical area and then validating them in another. To start improving classifier performance across the board, we first need to try to pinpoint the causes and scenarios in which the neural network does not perform well. In this sense, we can distinguish two different cases: 
\begin{itemize}
    \item network performance drops when it is applied to another geographical region in the \emph{same aerial picture}, and
    \item network performance drops when the network is applied to a \emph{different aerial picture}.
\end{itemize}

When network performance drops in another geographical region of the same map, the network has been overfitted to one specific geographical area. In this case, the differences between both geographical areas have to be analyzed. It could be the case that there are differences between rural and more urban areas and also the architecture may vary from city to city. Therefore, we would like to take trained networks and further evaluate them in different urban settings. In specific, geographical areas that are further apart can have larger differences in building style, architecture, and urban planning. By comparing the correctly identified cases and those which are not, we can get an idea which features the network uses to classify the images. By using visualization techniques like class activation maps and Deep Taylor Decomposition \citep{DBLP:journals/corr/SelvarajuDVCPB16, zhou2015cnnlocalization, DBLP:journals/corr/MontavonBBSM15}, it is possible to identify which parts of the image the network uses to come to its classification. By especially looking at the images that were falsely identified, a cause for dropping classifier performance may be identified. In addition, we would like to investigate whether a more realistic sampling of the training set would lead to better performance in this situation.

When network performance drops when the network is applied to another aerial picture, the causes can be even more varied. Different aerial images can contain the same geographical area, but can vary in resolution or for example the time period/year in which the photo was taken. Aerial images from different years can be flown in different conditions. The aerial image can for instance be taken during another time of day, in different weather, or with different camera equipment. Slightly different flight paths can also result in other camera angles for the same house. Moreover, image normalization may vary from year to year. Conversely, different aerial images may also depict geographical areas that are completely different, like for example different countries, with again different urban planning, building styles, and architectures. By carefully comparing network performance in these varying conditions, we can identify the conditions the network is most sensitive to. Is network performance mostly influenced by differences in resolution, illumination and image normalization, by different camera angles, by differences in background, or are differences in architecture and urban planning more important in the end? By trying to pinpoint the reasons of deteriorating network performance we can try to improve classifier performance. 

\section{Improving network performance}
\label{section:improving_classifier_performance}
To improve network performance there are several options. Of course, which options will be used also largely depends on what was found to cause deteriorating network performance (see section \ref{section:causes_classifier_performance}). To improve network performance, we have roughly three options:
\begin{itemize}
    \item Improve the dataset,
    \item Improve the model,
    \item Improve the training process.
\end{itemize}

To \emph{improve the dataset} we could either add more data to the dataset or sample the data that we use in a smarter way. Currently, we sampled specific geographical areas and trained the network on those areas. We sampled the built-up area of a certain city (Heerlen) or performed a random sample of a certain geographical area (Northrhein Westphalia). Both sampling methods may result in certain biases in the dataset, which result in lower model performance on a much larger geographical area. Therefore, a smarter sampling method is needed that reflects the class distributions and variation present in the target population. Such a sample can be made using, for example, building registers to sample on the basis of building year, building type, and even to provide a dataset that provides enough "architectural variation". Likewise, land use registers can be used to create a training set that also provides enough variation in terms of background; by using land use information, we can filter the geographical regions down to urban areas, but also provide a balanced sampling of different types of geographical areas; such as forests, agricultural, and built-up areas. From a preliminary look at the class activation maps, it becomes clear that for the negative cases the network trained on the Heerlen data is especially taking into account vegetation. Something similar may have happened to the network trained on the NRW data. That may explain the network performing better on a more rural city like Düren than a really urbanized area like Bonn. 

What is more, when sampling large geographical areas we need to make sure that every part of the geographical areas is represented accordingly. In addition to that, we could look at the cases the network has trouble identifying correctly, and over-sample those cases in the training set. Especially, the false positives and false negatives are important in this respect, but also cases that lead to a less clear-cut network decision. Moreover, it is not only important to sample the training set smartly, but also the test set and validation set should contain enough variation and similar class distributions so network performance reflects the performance in the real world. 

Of course, in contrast to smart sampling, another way to increase variation is just increasing the size of the training set. At this moment, our training sets had a size of around 15000 images, making the training sets small, especially compared to the DeepSolar project \citep{1136}, that classified solar panels across the US and used 360000 images. Specifically, we think that differences between European countries in building style, architecture and urban planning are likely bigger than the differences in the US, which may indicate that we need an even bigger training set for the EU.

To \emph{improve the model} more network architectures can be evaluated, different types of regularization, fully connected layers, and activation functions. For the lower resolutions, we would like to evaluate custom architectures with a lower number of layers than the out-of-the-box networks used for ImageNet. Another way to increase network performance, is using ensembles of networks that each were trained in a slightly different way. This often leads to improvements in accuracy as big as 5\% \citep{rosebrock_dl4cv}.

To \emph{improve the training process} we can especially look at different optimizers, learning rate schedules, and image data augmentation. If it follows that networks have troubles with image color temperature, illumination, or normalization, we can look into custom types of image data augmentation especially varying these aspects of images. Another way would be to artificially create more training samples using techniques such as Generative Adversarial Networks \citep{NIPS2014_5423}.

\section{Minimizing the effects of class imbalance} 
\label{section:minimize_effect_class_imbalance}
The datasets used in the evaluations in this report were heavily imbalanced. The imbalance is caused by the number of buildings without solar panels largely outnumbering the number of buildings with solar panels. On the basis of the register data, for the part of the Netherlands we considered, the number of solar panels is estimated to be about 8\% (likely a bit more). To train a model that generalizes across geographical areas it is important that the sample from the dataset used for training, testing and validation reflects the real distribution of solar panels versus non solar panels. This distribution can vary largely from area to area. All of the metrics used in this report are more or less susceptible to class imbalance, and therefore may vary from area to area. For example, in a dataset where the negative cases are in the majority, the positive precision is heavily influenced by the imbalance: the probability for false positives is larger than with a dataset where the ratio between positives and negatives is more balanced. The positive precision will be therefore much lower for a dataset where the negative cases largely outnumber the positive cases. In this example, the positive F1-score being the harmonic average of the precision and recall is also highly influenced by the imbalance. Conversely, the positive recall, negative precision and recall are less susceptible to the data imbalance when there a more negative than positive cases. To compare network performance in a more reliable way across datasets with varying distributions of positives versus negatives, evaluation metrics addressing (and not influenced by) the imbalance should be considered in further work, see for example \citep{luque_impact_2019, jeni_facing_2013, amin_comparing_2016}.

\section{One network as unique solution}
\label{section:one_network}
As we have seen from the results in this report, training a deep learning network that can be applied in a cross-country context is challenging. A network that would be applicable to the whole European Union may prove to be a very ambitious task. We feel that it may be more realistic to train specialized networks that perform well in certain geographical areas. We think that the method and networks developed in this report may be a good starting point for such local classifiers. The weights learned on the aerial images above Netherlands and Germany can be used as a starting point for further fine-tuning.

\section{Creating a model zoo for aerial images}
\label{section:model_zoo}
Within the deep-learning community, it is customary for researchers to share model structures and pre-trained weights in so-called model zoos \footnote{\url{https://modelzoo.co/}} \footnote{\url{https://github.com/tensorflow/models/blob/master/research/object_detection/g3doc/detection_model_zoo.md}} \footnote{\url{https://caffe.berkeleyvision.org/model_zoo.html}}. Model zoos facilitate transfer learning and can greatly reduce the effort to train neural networks for certain domains. To our knowledge, no model zoo has been created specifically for aerial images. To facilitate other researchers using aerial images and make it easier for them to start similar projects, we would like to share the results of the DeepSolaris project and create a model zoo specifically for aerial images. Such a model zoo could also be a platform to share models for solar panel detection across the European countries.

\section{Developing a cross-border dataset}
\label{section:cross_border_dataset}

In addition to sharing our models in a model zoo, we would also like to share a pre-annotated, cross-border dataset for the province of Limburg and North Rhine Westphalia. At first, we will provide the high resolution DOP10 images for both countries, adding lower resolutions at a later stage. A cross-border dataset provides a good benchmark for other projects in the area, or even in other European regions. Other than that, a pre-labeled dataset provides a good starting point for other researchers to start training their own models.

\section{Comparing with Register Data}
\label{section:comparison_with_register_data}
Finally, we would like to compare the results of the network predictions for Limburg and NRW with the register data available. By comparing the predictions with the labels from the registers we can improve the quality of these registers. First of all, we found that the registers for both Limburg and NRW contained many false positives; the register indicated that a solar panel was present at a certain location, but a visual inspection did not reveal any. By comparing results, the false positives can be removed from the registers. In addition, several solar panels were found that were not available in the registers yet. By adding these solar panels, we can further improve register quality but also give an estimate of how many installations are missing. This may be an important estimate when looking into other geographical regions as well.

To improve register quality in a semi-automatic way, we would moreover like to look into active learning techniques \citep{Bernard2018a, Babaee2015, Settles2009}, in which the model training and annotation process are intermingled, and model results are checked by a human annotator. In this way, only those results have to be checked for which the model is uncertain, or the register data does not match. The amount of data manually checked can be greatly reduced by using such an approach.

An important point to consider when comparing model predictions with the register data is the geo-localisation of the addresses of the buildings predicted to have solar panels. The classification models we used in this report give a single label (solar panel/no solar panel) prediction for a tile of a certain dimension, for instance $20\times20m$. One tile can contain several buildings and as such solar panel addresses can only reliably be derived for tiles containing a single building. To be able to derive precise information about those buildings containing solar panels, the object detection models, also discussed in this report, need to be used. The object detection models give the shape and location of solar panels which can be assigned to a building and address by combining the model predictions with the building shapes and addresses from a building register.

\section*{Acknowledgements}
This research is being conducted under the ESS action 'Merging Geostatistics and Geospatial Information in Member States' (grant agreement no.: 08143.2017.001-2017.408) and a CBS investment for the development of a Deep Learning algorithm.
\newpage

\bibliographystyle{plain}
\bibliography{Bibliography.bib}

\appendix
\section{Appendix}
\subsection{Source Code}
The Python source code to this project is published under an open source MIT-based licence. The source code developed at Statistic Netherlands, is available at \url{https://www.gitlab.com/CBDS/DeepSolaris}. This repository contains code to perform feature extraction with deep learning models, fine-tune and evaluate the deep learning networks, and several notebooks containing explorative studies. Another repository \url{https://gitlab.com/CBDS/MapTiler} contains the software, developed in C++, used to create the tiles from the aerial images. This code was published under an LGPLv3 licence. The code developed by the Open University, can be found here: \url{https://github.com/SB-BISS/DeepSolarisObjectDetection }.

\section{Glossary}
\newcommand{\specialcell}[3][c]{%
    \begin{tabular}[#1]{@{}#2@{}}#3\end{tabular}}

 \begin{longtable}{|l|p{8cm}|}
 \hline
  \textbf{ } & \textbf{Definition}\\
 \hline
 \endfirsthead

 \hline
\textbf{ } & \textbf{Definition}\\
 \hline
 \endhead

  \hline
  Aerial image & A photograph from an aircraft or other flying object.   \\
  \hline
  Annotation & Annotation is a process of manually labelling data on images containing specific objects to make it recognizable for machines.  \\
  \hline
  Classification algorithm &  Classification of different objects which could  flowers, faces,  fruits or any object we could  imagine. \\
  \hline
  CIR & Color-Infrared. \\
  \hline
  \specialcell{l}{Convolutional Neural Networks \\ (CNN)}  & A convolutional neural network (CNN) is an artificial neural network. It is a concept inspired  by biological processes. Convolutional Neural Networks are used in numerous modern  artificial intelligence technologies, primarily for the machine processing of image or audio data. \\
  \hline
  DOP25 and DOP10 & A digital orthophoto is an aerial photograph or a satellite imagery geometrically corrected. This rater dataset has been accurately scanned and rectified with the aid of geodetic surveying and photogrammetry. Digital orthophotos are available in different scales (DOP10, 1:1000; DOP25, 1:2500). \\
   \hline
   DOP10 WMS-service & A Web Map Service (WMS) is a web-based map service used to retrieve excerpts from maps via the Internet.\\
  \hline
  F1-score & Measure of accuracy which is useful when you want to seek a balance between precision and recall. \\
  \hline
  Keras & Keras is an open-source neural –network library written in Python.  \\
  \hline
  LANUV registry & LANUV is the abbreviation for the North Rhine-Westphalia State Office for Nature, the Environment and Consumer Protection (Landesamt für Natur, Umwelt und Verbraucherschutz Nordrhein-Westfalen). The LANUV dataset referred to in this paper is a dataset containing the registered solar panels.\\
  \hline
 NSOs & National Statistical Offices \\
 \hline
 Object detection & Object detection is a computer technology related to computer vision and image processing that deals with detecting instances of semantic objects of a certain class (such as humans, buildings, or cars) in digital images and videos. \\
 \hline
 Polygon & A polygon is a plane figure that is described by a finite number of straight line segments connected to form a closed polygonal chain or polygonal circuit. \\
 \hline
 Resolution & In the context of remote sensing there are four different types of resolution: spatial, spectral, radiometric, and temporal. The spatial resolution describes the area represented by each pixel of an image. The temporal resolution concerns the time laps between two successive images of the same target area. Spectral resolution refers to the resolving power of a system in terms of wavelength or frequency. Radiometric resolution refers to the resolving power of a system in terms of the signal energy (detection of energy differences (reflection and emission) in terms of temperature, intensity and power). \\
 \hline
 Sigmoid & A sigmoid function is an activation function, which bounds the output between 0 and 1. \\
\hline
SVM classifier & Support-vector machines are supervised learning models with associated learning algorithms for classification and regression analysis in the context of machine learning. \\
\hline
Scikit-learn & a Python package containing all kinds of machine learning tools, ranging from classification algorithms, to regression and clustering algorithms, but also containing a lot of helper functions, for instance for the evaluation of machine learning performance. See \url{https://scikit-learn.org} \\
\hline
\end{longtable}

\end{document}